\documentclass[sigconf, nonacm]{acmart}
\usepackage[normalem]{ulem}     
\useunder{\uline}{\ul}{}        
\usepackage{amsmath} 
\usepackage[ruled,lined]{algorithm2e}
\usepackage{caption} 
\usepackage{subfigure}
\usepackage{bm}               
\usepackage{enumitem}

\usepackage{xcolor}
\usepackage{colortbl}
\usepackage{float}
\usepackage{placeins}

\definecolor{mypink}{rgb}{0.95, 0.5, 0.7}
\definecolor{myblue}{rgb}{0.0, 0.5, 0.8}
\definecolor{darkbrown}{rgb}{0.58, 0.81, 0.58}
\definecolor{lightbrown}{rgb}{0.69, 0.79, 0.89}
\definecolor{darkblue}{rgb}{0.0, 0.7, 0.7}
\definecolor{lightblue}{rgb}{0.0, 0.0, 0.25}
\definecolor{visualization_yellow}{rgb}{0.9, 0.6235, 0.2470}
\definecolor{visualization_green}{rgb}{0.5921, 0.81569, 0.59216}

\AtBeginDocument{%
  }

\setcopyright{acmlicensed}
\copyrightyear{2018}
\acmYear{2018}
\acmDOI{XXXXXXX.XXXXXXX}

\acmConference[Conference acronym 'XX]{Make sure to enter the correct
  conference title from your rights confirmation emai}{June 03--05,
  2018}{Woodstock, NY}
\acmISBN{978-1-4503-XXXX-X/18/06}

\begin{document}

\title{HC-GLAD: Dual Hyperbolic Contrastive Learning for Unsupervised Graph-Level Anomaly Detection}

\author{Yali Fu}
\authornote{Equal contributions.}
\affiliation{
  \institution{Jilin University}
  \city{Changchun}
  \country{China}
}
\email{fuyl23@mails.jlu.edu.cn}

\author{Jindong Li}
\authornotemark[1]
\affiliation{
  \institution{Jilin University}
  \city{Changchun}
  \country{China}}
\email{jdli21@mails.jlu.edu.cn}

\author{Jiahong Liu}
\affiliation{
  \institution{The Chinese University of Hong Kong}
  \city{Hong Kong SAR}
  \country{China}
}
\email{jiahong.liu21@gmail.com}

\author{Qianli Xing}
\affiliation{
 \institution{Jilin University}
 \city{Changchun}
 \country{China}}
\email{qianlixing@jlu.edu.cn}

\author{Qi Wang}
\authornote{Corresponding author.}
\affiliation{
  \institution{Jilin University}
  \city{Changchun}
  \country{China}\\
  \institution{Engineering Research Center of Knowledge-Driven Human-Machine Intelligence, Ministry of Education, China}
}
\email{qiwang@jlu.edu.cn}

\author{Irwin King}
\affiliation{
  \institution{The Chinese University of Hong Kong}
  \city{Hong Kong SAR}
  \country{China}}
\email{king@cse.cuhk.edu.hk}


\begin{abstract}
Unsupervised graph-level anomaly detection (UGAD) has garnered increasing attention in recent years due to its significance. Most existing methods that rely on traditional GNNs mainly consider pairwise relationships between first-order neighbors, which is insufficient to capture the complex high-order dependencies  often associated with anomalies. This limitation underscores the necessity of exploring high-order node interactions in UGAD. In addition, most previous works ignore the underlying properties (e.g., hierarchy and power-law structure) which are common in real-world graph datasets and therefore are indispensable factors in the UGAD task. In this paper, we propose a novel Dual \textbf{H}yperbolic \textbf{C}ontrastive Learning for Unsupervised \textbf{G}raph-\textbf{L}evel \textbf{A}nomaly \textbf{D}etection (HC-GLAD in short). To exploit high-order node group information, we construct hypergraphs based on pre-designed gold motifs and subsequently perform hypergraph convolution. Furthermore, to preserve the hierarchy of real-world graphs, we introduce hyperbolic geometry into this field and conduct both graph and hypergraph embedding learning in hyperbolic space with the hyperboloid model. To the best of our knowledge, this is the first work to simultaneously apply hypergraph with node group information and hyperbolic geometry in this field. Extensive experiments on 13 real-world datasets of different fields demonstrate the superiority of HC-GLAD on the UGAD task. The code is available at \url{https://github.com/Yali-F/HC-GLAD}.
\end{abstract}



\maketitle

\section{Introduction}
\begin{figure}
    \centering
    \subfigure[High-order information in network data for toy examples.]{\includegraphics[width=0.45\textwidth]
    {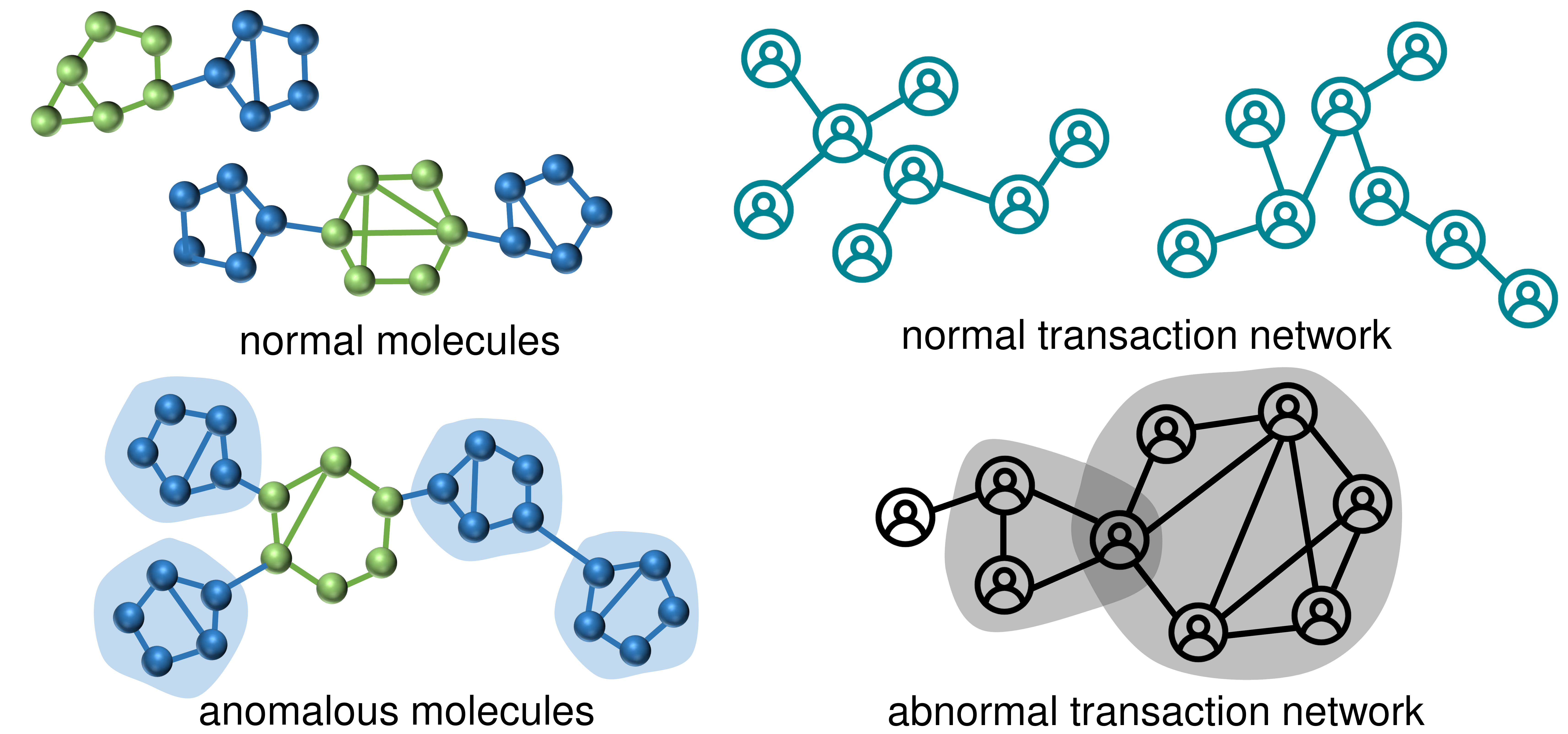}\label{fig:fig_1_molecules}}
    \hspace{0.1mm}
    \subfigure[Characters of two space.]{\includegraphics[width=0.45\textwidth]{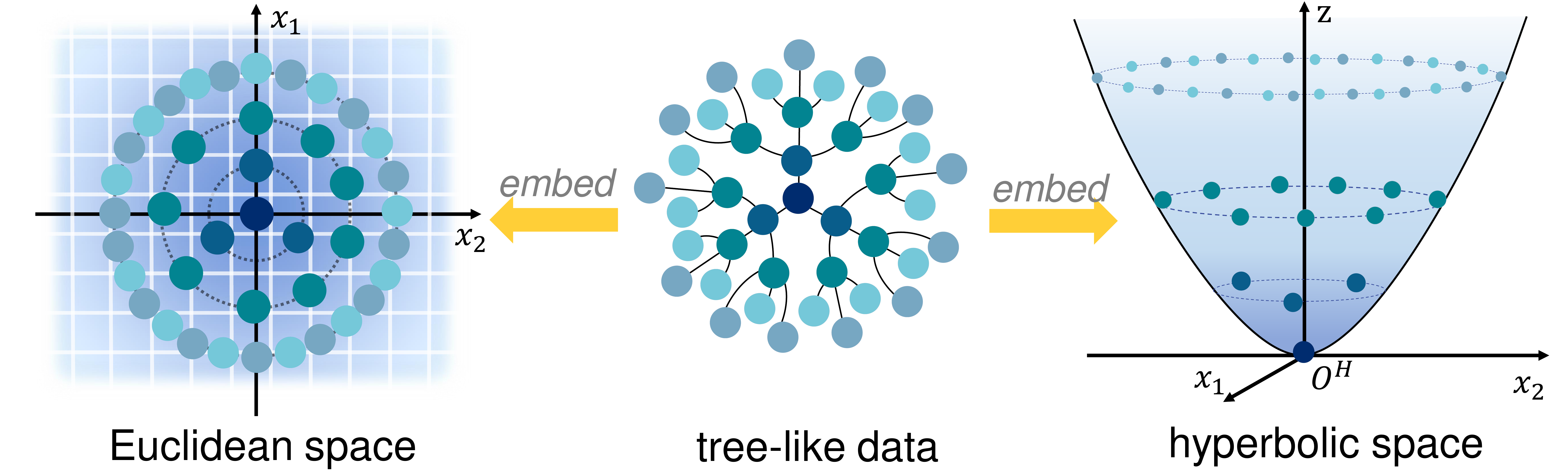}\label{fig:fig_1_Hyperbolic-Euclidean}}
    \caption{(a) Normal molecular graphs usually have 1-2 node groups in blue areas, while abnormal ones have 3-4. Normal financial transaction networks show simple patterns, while abnormal ones show chaotic circular or cross-transactions in gray areas; (b) With an exponential increase of nodes in tree-like data, Euclidean space is difficult to embed nodes separately. In contrast, hyperbolic space, which can be regarded as a continuous version of the tree, can still maintain certain distances between the embedded nodes.}
    \label{fig:fig_1_3}
\end{figure}

Graph-level anomaly detection helps uncover anomalous behaviors hidden within complex graph structures, which has been widely applied in various fields, including social network analysis, bioinformatics, and network security~\cite{2022_WSDM_GLocalKD, 2022_ScientificReports_GLADC, 2023(2024)_NeurIPS_SIGNET}. Unlike traditional anomaly detection methods that focus on individual data points or samples, graph-level anomaly detection focuses on the overall structure, topology, or features of the entire graph. Recently, there has been a growing interest in unsupervised graph-level anomaly detection (UGAD) as it offers an advantage by not relying on labeled data, rendering it applicable across various real-world scenarios. 
Despite the considerable research and exploration already conducted in this area \cite{2023_WSDM_GOOD-D, 2023_DASFAA_TUAF, 2023_ECMLPKDD_CVTGAD, 2023_ECMLPKDD_HimNet}, there are still several issues that need to be further explored.

\textbf{Firstly}, most existing UGAD methods using GNNs treat edges and nodes as fundamental units for message passing~\cite{2023_DASFAA_TUAF}, relying solely on pairwise relationships to capture key patterns. However, real-world networks often involve group interactions among multiple nodes, particularly in protein-protein interactions (PPI), molecular complexes, and delocalized bonds among multiple atoms~\cite{2024_hyperGraph_MHNN, 2021_hyperGraph_MolHMPN}. In some cases, anomalous graphs arise from these complex interactions among multiple nodes, exhibiting patterns significantly different from normal graphs. 
As shown in Figure~\ref{fig:fig_1_molecules}, the decisive factor in determining whether a molecule is anomalous lies in distinct groups connected to the central group: while normal molecular graphs typically involve fewer external groups, anomalous ones exhibit more complex group structures~\cite{2023(2024)_NeurIPS_SIGNET}. Similarly, abnormal financial transactions are characterized by more intricate transaction patterns, where anomalies often stem from irregular or convoluted interactions among multiple accounts, differing from the straightforward patterns in normal transactions.
There is an urgent need to holistically consider such complex group interactions to capture key patterns for anomaly detection.

\begin{figure}
    \centering
    \includegraphics[width=0.21\textwidth]{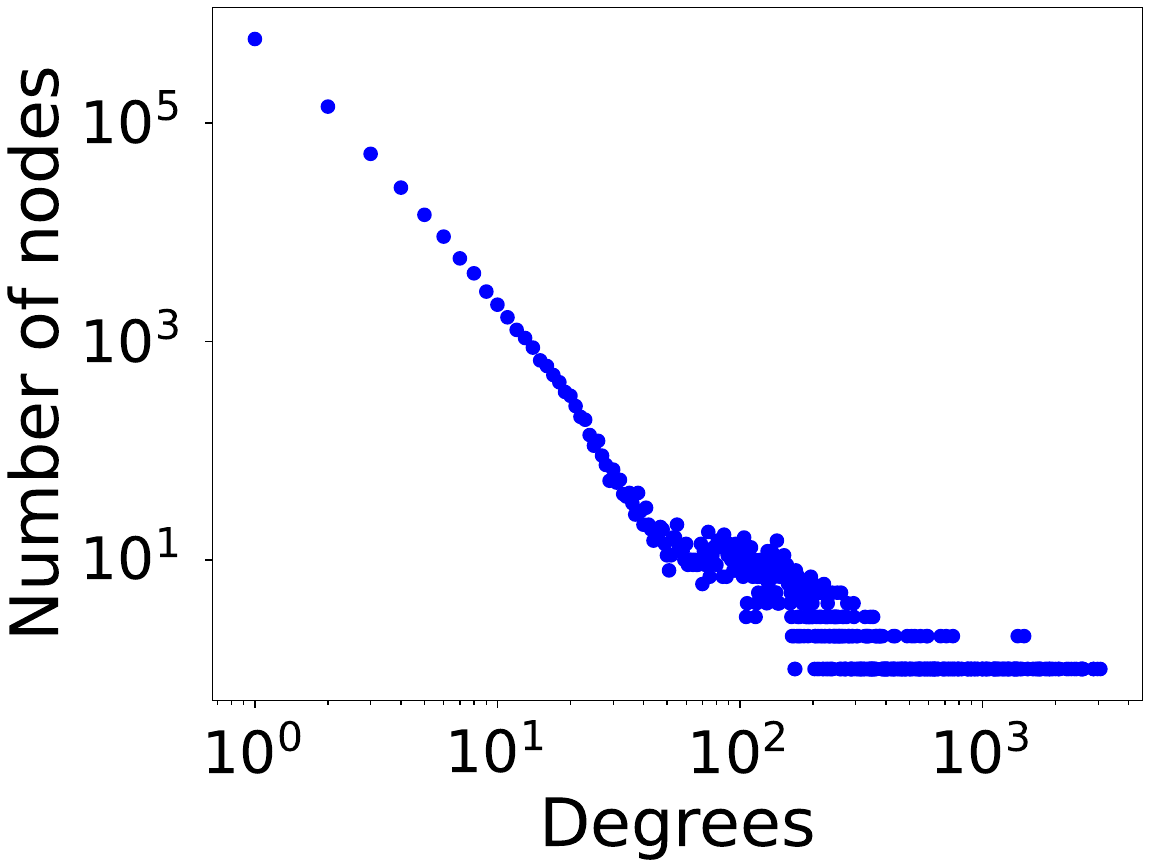}
    \hspace{2mm}
    \includegraphics[width=0.21\textwidth]{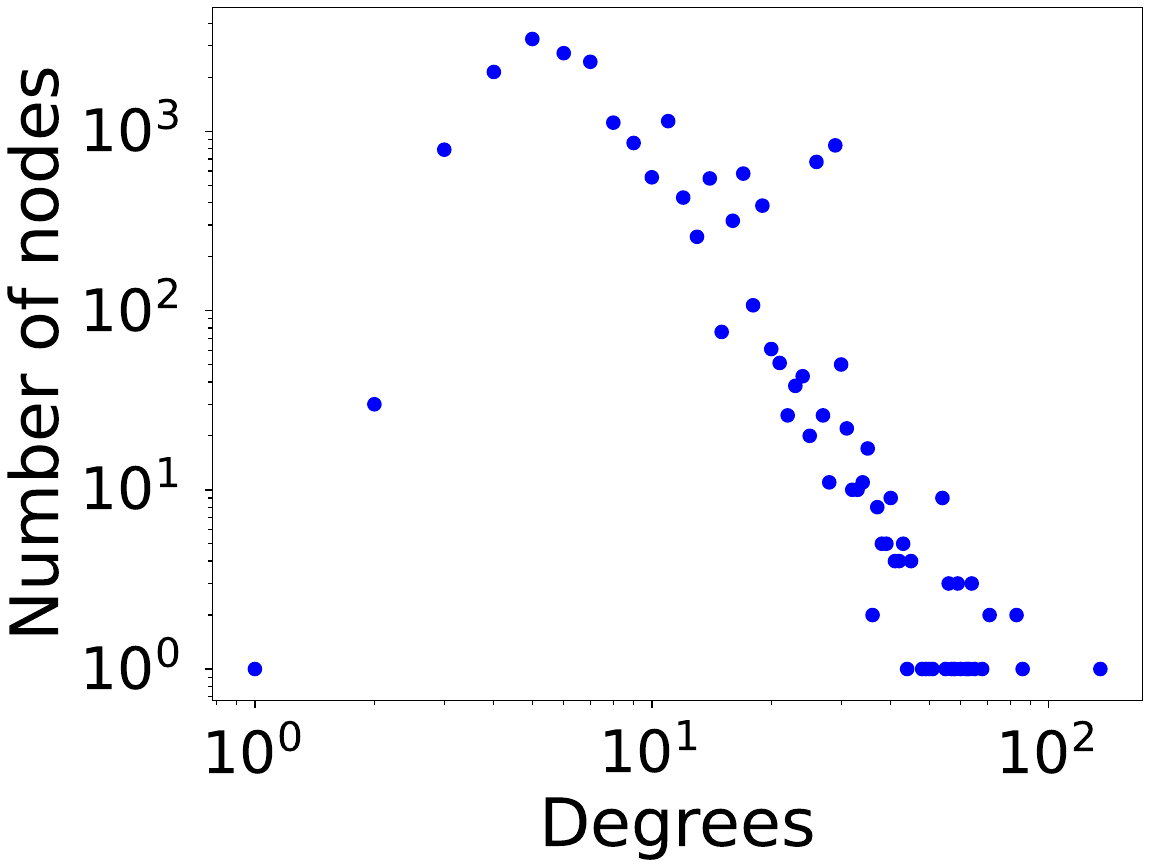}
    \caption{Degree distributions of dataset REDDIT-B (on the left) and IMDB-B (on the right).}
    \label{fig:fig_2}
\vspace{-1.5em}
\end{figure}

\textbf{Secondly}, the majority of current methods are based on GNNs established in Euclidean space~\cite{2022_WSDM_GLocalKD, 2023_WSDM_GOOD-D}, but the dimensionality of Euclidean space brings a fundamental limitation on its ability to represent complex networks~\cite{2014_NeurIPS_ARE}. It has been demonstrated that numerous real-world datasets exhibit characteristics akin to those of complex networks, including degree power-law distribution as shown in Figure~\ref{fig:fig_2}, which embodies latent tree-like hierarchical structure \cite{2003_PhysicalReview_Hierarchical, 2020_ICML_WHC_Hyperbolic}. In such tree-like data, the number of nodes has an exponential growth trend. For instance, the total number of nodes is $2^{(h+1)}-1$ ($h$ is the height of the tree) in a full binary tree. Nevertheless, flat Euclidean space whose volume grows polynomially cannot embed latent hierarchies without suffering from high distortion, as shown in Figure~\ref{fig:fig_1_Hyperbolic-Euclidean}. Therefore, it is necessary to employ a new paradigm or space to exploit the latent hierarchical information in UGAD.

Based on the aforementioned challenges and analysis, we propose a novel Dual \textbf{H}yperbolic \textbf{C}ontrastive Learning for Unsupervised \textbf{G}raph-\textbf{L}evel \textbf{A}nomaly \textbf{D}etection framework, namely HC-GLAD. 
In concrete, for the first challenge, we introduce hypergraph to naturally capture high-order structures beyond pairwise relationships. The anomaly-aware hypergraph is constructed based on pre-designed gold motifs. Compared to traditional graph structures, hyperedges in hypergraphs can connect multiple nodes, and hypergraph convolution enables message propagation in a broader context, yielding more comprehensive feature representations. This approach can not only capture local node interactions but also integrate global information through hyperedges, identifying high-order structures that deviate from normal patterns, thereby improving the accuracy of anomaly detection. 
For the second challenge, we incorporate hyperbolic geometry into UGAD. 
The curved hyperbolic space can be seen as a continuous version of the tree with exponential growth~\cite{2022_WWW_HRCF_Hyperbolic}, allowing it to naturally retain the rich hierarchical information in graph data. Also, under the same radius, the hyperbolic space can accommodate more nodes for informative embeddings as shown in Figure~\ref{fig:fig_1_Hyperbolic-Euclidean}.
Specifically, we utilize the hyperboloid model to preserve the latent hierarchical information and conduct both graph and hypergraph embedding  in hyperbolic space. Our major contributions are summarized as follows:
\begin{itemize}
    \item We propose a novel dual hyperbolic contrastive learning for the unsupervised graph-level anomaly detection framework (HC-GLAD). To the best of our knowledge, this is the first work to simultaneously introduce hypergraph exploiting node group information and hyperbolic geometry to the unsupervised graph-level anomaly detection task.
    \item We utilize hypergraphs to explore node group information based on pre-designed gold motifs. In addition, we employ hyperbolic geometry to leverage latent hierarchical information and accomplish achievements that cannot be attained in Euclidean space. The advantages of hypergraph learning, hyperbolic learning, and contrastive learning are integrated into a unified framework to jointly improve model performance. 
    \item We conduct extensive experiments on 13 real-world datasets, demonstrating the effectiveness and superiority of HC-GLAD for unsupervised graph-level anomaly detection.
\end{itemize}

\begin{figure*}
    \centering
    \includegraphics[width=0.99\textwidth]{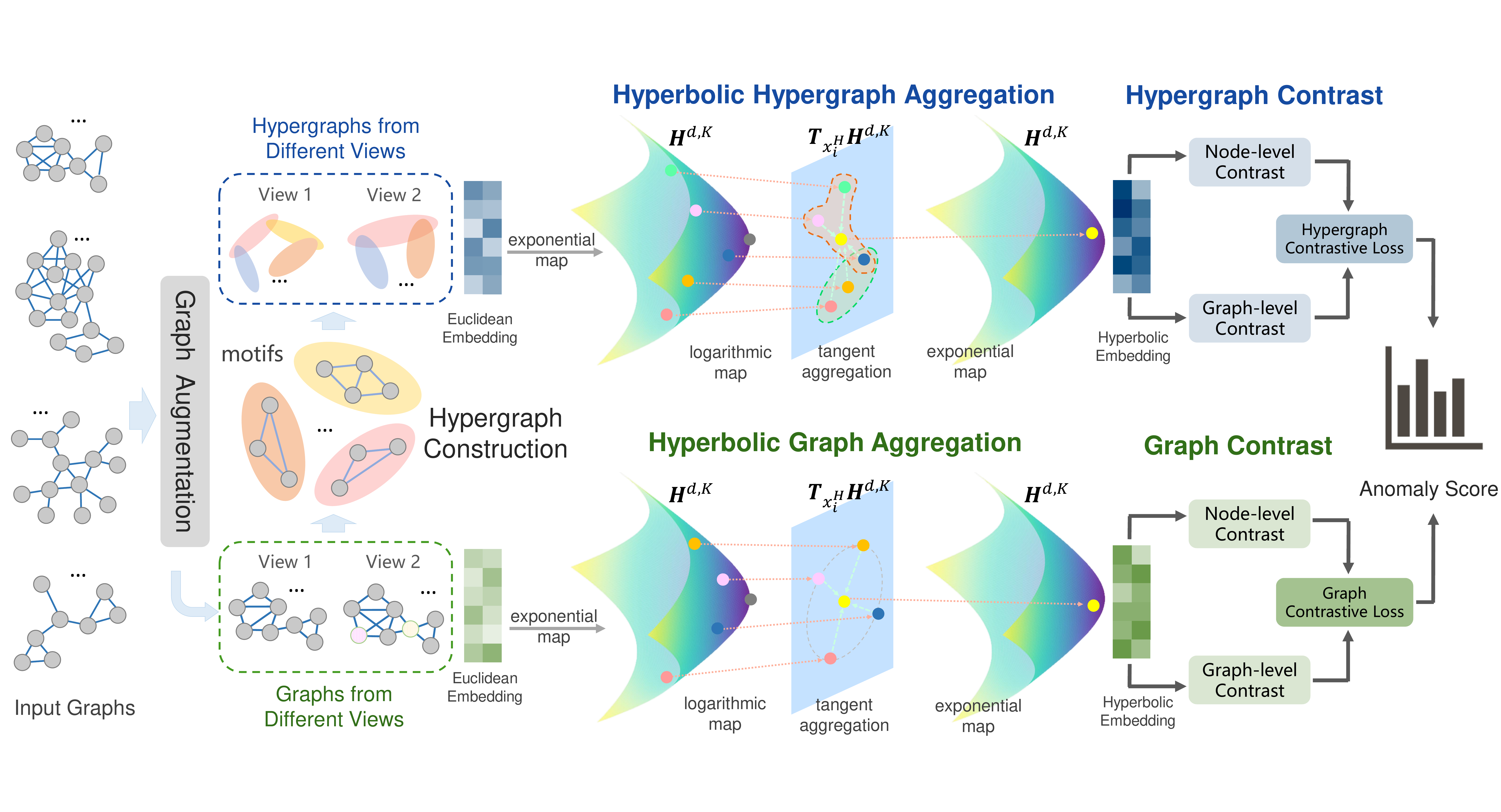}
    \caption{The overall framework of HC-GLAD. Firstly, the input graphs undergo data augmentation in Euclidean space, obtaining two augmented views and forming the graph channel below. Secondly, based on pre-designed gold motifs, we construct hypergraphs from the augmented graphs, forming the hypergraph channel above. Thirdly, we perform the aggregation operation for graph and hypergraph channel in hyperbolic space, obtaining final hyperbolic embeddings used for calculating multi-level contrastive loss. Lastly, graph and hypergraph contrastive losses are employed to calculate graphs' anomaly scores.}
    \label{fig: framework}
\vspace{-0.12em}
\end{figure*}

\section{Related Work}
\subsection{Graph-Level Anomaly Detection}
In the context of graph data analysis, the objective of graph-level anomaly detection is to discern abnormal graphs from normal ones, wherein anomalous graphs often signify a minority but pivotal patterns \cite{2021_TKDE_SurveyofGAD}. Nowadays, there are numerous methods that explore graph-level anomalies in graphs. OCGIN~\cite{2021(2023)_BigData_OCGIN} is the first representative model, and it integrates the one-class classification and graph isomorphism network (GIN) \cite{2019_ICLR_GIN} into the graph-level anomaly detection. OCGTL~\cite{2022_IJCAI_OCGTL} integrates the strengths of deep one-class classification and neural transformation learning. GLocalKD~\cite{2022_WSDM_GLocalKD} implements joint random distillation to detect both locally anomalous and globally anomalous graphs by training one graph neural network to predict another graph neural network. GOOD-D~\cite{2023_WSDM_GOOD-D} introduces perturbation-free graph data augmentation and performs hierarchical contrastive learning to detect anomalous graphs based on semantic inconsistency in different levels. TUAF~\cite{2023_DASFAA_TUAF} builds triple-unit graphs and further learns triple representations to simultaneously capture abundant information on edges and their corresponding nodes. CVTGAD~\cite{2023_ECMLPKDD_CVTGAD} applies transformer and cross-attention into UGAD, directly exploiting relationships across different views. SIGNET~\cite{2023(2024)_NeurIPS_SIGNET} proposes a multi-view subgraph information bottleneck framework and further infers anomaly scores and provides subgraph-level explanations.

\subsection{Hyperbolic Learning on Graphs}
Hyperbolic learning has attracted massive attention from the graph fields such as recommendation systems, node classification, and molecular learning due to its superior geometry property (i.e., its volume increases exponentially in proportion to its radius) of hyperbolic space compared to Euclidean space \cite{2022_arXiv_Survey_Hyperbolic, 2022_KDD_HICF}. HGNN (hyperbolic graph neural network)~\cite{2019_NeurIPS_HGNN_Hyperbolic} generalizes the graph neural networks to Riemannian manifolds and improves the performance of the full-graph classification task. It fully utilizes the power of hyperbolic geometry and demonstrates that hyperbolic representations are suitable for capturing high-level structural information. HGCN (hyperbolic graph convolutional neural network) \cite{2019_NeurIPS_HGCN_Hyperbolic} leverages both the expressiveness of GCNs and hyperbolic geometry.
$\kappa$-GCN \cite{2020_ICML_k-GCN} presents an innovative expansion of GCNs to encompass stereographic models with both positive and negative curvatures, thereby offering a unified approach. HAT (hyperbolic graph attention network) \cite{2021_BigData_HAT_Hyperbolic} proposes the hyperbolic multi-head attention mechanism to acquire robust node representation of graph in hyperbolic space and further improves the accuracy of node classification. 
LGCN \cite{2021_WWW_LGCN} introduces a unified framework of graph operations on the hyperboloid (i.e., feature transformation and non-linearity activation), and proposes an elegant hyperbolic neighborhood aggregation based on the centroid of Lorentzian distance. HRCF \cite{2022_WWW_HRCF_Hyperbolic} designs a geometric-aware hyperbolic regularizer to boost the optimization process by the root alignment and origin-aware penalty, and it enhances the performance of a hyperbolic-powered collaborative filtering. HyperIMBA~\cite{2023_WWW_HyperIMBA_Hyperbolic} explores the hierarchy-imbalance issue on hierarchical structure and captures the implicit hierarchy of graph nodes by hyperbolic geometry.

\subsection{Hypergraph Learning}
Due to the capability and flexibility in modeling complex correlations of graph data, hypergraph learning has earned more attention from both academia and industry \cite{2022_TPAMI_Survey_Hypergraph}. Hypergraphs naturally depict a wide array of systems characterized by group relationships among their interacting parts~\cite{2023_ACM_Survey_HyperGraph}. 
HGNN (hypergraph neural network) \cite{2019_AAAI_HGNN_Hypergraph} designs a hyperedge convolution operation and encodes high-order data correlation in a hypergraph structure. 
HyperGCN~\cite{2019_NeurIPS_HyperGCN} utilizes tools from the spectral theory of hypergraphs and introduces a novel way to train GCN for semi-supervised learning and combinatorial optimization tasks. 
HGNN$^+$ \cite{2022_TPAMI_HGNN+_Hypergraph} introduces "hyperedge groups" to capture high-order correlations in multi-modal data and uses an adaptive fusion strategy to integrate them into a unified hypergraph. This allows the model to better represent complex relationships across different data types.
DHCF \cite{2020_KDD_DHCF} constructs two hypergraphs (i.e., user and item hypergraph) and introduces a jump hypergraph convolution (jHConv) to enhance collaborative filtering recommendation performance. 
HHGR \cite{2021_CIKM_HHGR_Hypergraph} builds user-level and group-level hypergraphs and employs a hierarchical hypergraph convolution network to capture complex high-order relationships within and beyond groups, thus improving the performance of group recommendation. 
DH-HGCN \cite{2022_SIGIR_DH-HGCN_Hypergraph} utilizes both a hypergraph convolution network and homogeneity study to explicitly learn high-order relationships among items and users to enhance multiple social recommendation performance. 
HCCF \cite{2022_SIGIR_HCCF_Hypergraph} designs a hypergraph-enhanced cross-view contrastive learning architecture to jointly capture local and global collaborative relations in the recommender system. 

An extensive review of the literature is included in Appendix~\ref{appendix:gcl}.

\section{Preliminaries}
\subsection{Hyperboloid Manifold}
Hyperbolic space, defined by its constant negative curvature, diverges from the flatness of Euclidean geometry. The hyperboloid manifold is often favored for its numerical stability, making it a popular choice in hyperbolic geometry applications~\cite{2018_ICML_nickel_Hyperbolic_RSGD}.

\begin{definition}[Minkowski Inner Product]
The inner product $\langle\mathbf{x}, \mathbf{y}\rangle_{\mathcal{L}}$ for vectors $\mathbf{x}, \mathbf{y} \in \mathbb{R}^{d+1}$ is defined by the expression $\langle\mathbf{x}, \mathbf{y}\rangle_{\mathcal{L}} = -x_0 y_0 + \sum_{i=1}^d x_i y_i$.
\label{def: inner product}
\end{definition}

\begin{definition}[Hyperboloid Manifold]
A $d$-dimensional hyperboloid manifold, denoted as $\mathcal{L}^d$, with a constant negative curvature, is defined as the Riemannian manifold $(\mathbb{H}^d, g_{\ell})$. Here, we adopt the constant negative curvature of $-1$, and $g_{\ell}$ is the metric tensor represented by $\operatorname{diag}([-1,1, \ldots, 1])$, and $\mathbb{H}^d$ is the set of all vectors $\mathbf{x} \in \mathbb{R}^{d+1}$ satisfying $\langle\mathbf{x}, \mathbf{x}\rangle_{\mathcal{L}} = -1$ and $x_0 > 0$.
\end{definition}
Next, the corresponding intrinsic distance function for two points $\mathbf{x}, \mathbf{y}\in \mathcal{L}^d$ is provided as:
\begin{equation}
    d_\mathcal{L}(\mathbf{x}, \mathbf{y}) = ~\mbox{arcosh}~(-\langle \mathbf{x}, \mathbf{y}\rangle _\mathcal{L}).
    \label{equ: Lorentzian_distance}
\end{equation}

\begin{definition}[Tangent Space]
For a point $\mathbf{x} \in \mathcal{L}^d$, the tangent space $\mathcal{T}_{\mathbf{x}} \mathcal{L}^d$ consists of all vectors $\mathbf{v}$ that are orthogonal to $\mathbf{x}$ under the Minkowski inner product. This orthogonality is defined such that $\langle\mathbf{x}, \mathbf{v}\rangle_{\mathcal{L}}=0$. Therefore, the tangent space can be expressed as:
$\mathcal{T}_{\mathbf{x}} \mathcal{L}^d=\left\{\mathbf{v}:\langle\mathbf{x}, \mathbf{v}\rangle_{\mathcal{L}}=0\right\}.$  
\end{definition}

\begin{definition}[Exponential and Logarithmic Maps]
\label{def: exp and log}
Let $\mathbf{x} \in \mathcal{L}^d$ and \( \mathbf{v} \in \mathcal{T}_{\mathbf{x}} \mathcal{L}^d \). The exponential map \( \exp_{\mathbf{x}}: \mathcal{T}_{\mathbf{x}} \mathcal{L}^d \rightarrow \mathcal{L}^d \) and the logarithmic map \( \log_{\mathbf{x}}: \mathcal{L}^d \rightarrow \mathcal{T}_{\mathbf{x}} \mathcal{L}^d \) are defined as follows:
\begin{equation}
    \exp_{\mathbf{x}}(\mathbf{v}) = \cosh (\|\mathbf{v}\|_{\mathcal{L}}) \mathbf{x} + \sinh (\|\mathbf{v}\|_{\mathcal{L}}) \frac{\mathbf{v}}{\|\mathbf{v}\|_{\mathcal{L}}}, 
\end{equation}
\begin{equation}
    \log_{\mathbf{x}}(\mathbf{y}) = d_{\mathcal{L}}(\mathbf{x}, \mathbf{y}) \frac{\mathbf{y} + \langle \mathbf{x}, \mathbf{y} \rangle_{\mathcal{L}} \mathbf{x}}{\left\|\mathbf{y} + \langle \mathbf{x}, \mathbf{y} \rangle_{\mathcal{L}} \mathbf{x}\right\|_{\mathcal{L}}},
\end{equation}
where \( \|\mathbf{v}\|_{\mathcal{L}} = \sqrt{\langle \mathbf{v}, \mathbf{v} \rangle_{\mathcal{L}}} \) denotes the norm of \( \mathbf{v} \) in \( \mathcal{T}_{\mathbf{x}} \mathcal{L}^d \).
\end{definition}

For computational convenience, the origin of the hyperboloid manifold denoted as \( \mathbf{o} = (1, 0, 0, \ldots, 0) \) in \( \mathcal{L}^d \), is selected as the reference point for the exponential and logarithmic maps. This choice allows for simplified expressions of these mappings.
\begin{equation}
\begin{aligned}
    \exp_{\mathbf{o}}(\mathbf{v}) &= \exp_{\mathbf{o}}\left(\left[0, \mathbf{v}^E\right]\right) \\
    &= \left(\cosh \left(\|\mathbf{v}^E\|_2\right), \sinh\left(\|\mathbf{v}^E\|_2\right)\frac{\mathbf{v}^E}{\|\mathbf{v}^E\|_2}\right),
\end{aligned}
\label{equ: exp0}
\end{equation}
where the $(,)$ denotes concatenation and the $\cdot^E$ denotes the embedding in Euclidean space~\cite{2021_WWW_LGCN}.

\subsection{Notations and Problem Statement}
\textbf{Notations.}
We denote a graph as $G = (\mathcal{V}, \mathcal{E})$, where $\mathcal{V}$ is the set of nodes and $\mathcal{E}$ is the set of edges. The topology (i.e., structure) information of $G$ is represented by adjacency matrix $A \in \mathbb{R}^{n \times n}$, where $n$ is the number of nodes. $A_{i, j} = 1$ if there is an edge between node $v_i$ and $v_j$, otherwise, $A_{i, j} = 0$. We denote an attributed graph as $G = (\mathcal{V}, \mathcal{E}, \mathcal{X})$, where $\mathcal{X} \in \mathbb{R}^{n \times {d_{attr}}}$ represents the feature matrix of node features. Each row of $\mathcal{X}$ represents a node's feature vector with $d_{attr}$ dimension. The graph set is denoted as $\mathcal{G} = \{G_1, G_2, ..., G_m \}$, where $m$ is the number of graphs in $\mathcal{G}$.

\textbf{Problem Statement.}
In this work, we focus on the unsupervised graph-level anomaly detection task: in the training phase, we train the model only using normal graphs; in the inference phase, given a graph set $\mathcal{G}$ containing normal graphs and anomalous graphs, HC-GLAD aims to distinguish anomalous graphs that are significantly different from normal graphs according to the anomaly score.

\section{Methodology}
In this section, we will introduce the dual hyperbolic contrastive learning for the unsupervised graph-level anomaly detection framework (namely, HC-GLAD). The overall framework and brief procedure are illustrated in Figure \ref{fig: framework}, and the pseudo-code is summarized in Algorithm~\ref{alg:algorithm}. 

\subsection{Data Preprocessing}
\textbf{Graph Data Augmentation.}
We employ the perturbation-free graph augmentation strategy \cite{2023_WSDM_GOOD-D, 2023_AAAI_FedStar} to generate two augmented views (i.e., $v_{1}$ and $v_{2}$) for an input graph $G$. Concretely, $v_1$ focuses more on attribute and is directly built by integrating the node attribute $\mathcal{X}$ (for attributed graph) and adjacency matrix $A$. $v_2$ focuses more on structure and is built by structural encodings from the graph topology and then it is combined with adjacency matrix $A$.

\textbf{Hypergraph Construction.}
After obtaining two augmented views of a graph, we essentially have two augmented graphs. Inspired by \cite{2020_VLDB_MoCHy, 2021_WWW_MHCN}, we leverage ternary relationships between nodes, named the "gold motif" (i.e., the triangular relationships formed by three nodes), which is fundamental and ubiquitous in network structures, to initially construct hypergraphs. Given the adjacency matrix $A$ of an augmented graph, we first construct the relationship matrix $A_{relation}$ of the constructed hypergraph by using the gold motifs. It can be calculated by:
\begin{equation}
    A_{relation} = (AA^{T}) \odot A = (AA) \odot A ,
\label{Eq: motifs}
\end{equation}
where $A^{T} = A$ beacause graph $G$ is an undirected graph and $A$ is symmetric. 

We determine the high-order relationships between vertices based on the matrix $\hat{A}_{realation} = A_{relation} + I_{N}$, where $I_{N}$ is the identity matrix. We further build the incidence matrix $\mathbf{H_{inc}}$, where if vertex $v_i$ is connected by hyperedge $\epsilon$, $H_{inc (i \epsilon)} = 1$, otherwise 0. While thoroughly investigating and utilizing the gold motifs, we must also consider instances that do not constitute this kind of high-order relationship and ensure the integrity of the entire graph. Therefore, we will also include the edges that are not part of the high-order relationships in the incidence matrix $\mathbf{H}_{inc}$. Finally, we get a hypergraph $HyperG$ with $N$ vertices and $M$ hyperedges. The high-order relationships in hypergraph $HyperG$ could be simply represented by the incidence matrix $\mathbf{H}_{inc} \in \mathbb{R}^{N\times M}$.

\subsection{Hyperbolic (Hyper-)Graph Convolution}
Before we conduct hyperbolic (hyper-)graph convolution, we insert a value 0 in the zeroth dimension of the Euclidean state of the node for both view $v_1$ and $v_2$. Refer to Eq. (\ref{equ: exp0}), the initial hyperbolic node state $\mathbf{e}^0$ could be obtained by:
\begin{equation}
    e^0_i = \exp_{\mathbf{o}} ([0, \mathbf{x}_i]),
\label{Equation: initial_hyperbolic_node_state}
\end{equation}
where $\mathbf{x}$ is the initial feature (or encoding) from augmented graphs (i.e., $v_1$ and $v_2$). $[0, \mathbf{x}]$ denotes the operation of inserting the value 0 into the zeroth dimension of $\mathbf{x}$ so that $[0, \mathbf{x}]$ could always be in the tangent space of origin \cite{2022_KDD_HICF, 2021_WWW_HGCF}. The superscript 0 in $e^0_i$ indicates the initial hyperbolic state.

\subsubsection{Hyperbolic Graph Aggregation}
Following \cite{2022_KDD_HICF, 2021_WWW_HGCF}, we first map the initial embedding $e^0_i$ in hyperbolic space to the tangent space using the logarithmic map.
Then, we select GCN as our fundamental graph encoder to perform graph convolution aggregation. The propagation rule in the $l$-th layer on the view $v_1$ can be expressed as:
\begin{equation}
    \mathbf{H}^{(v_1,~ l)}_{graph} = \sigma \left(\hat{\mathbf{D}}^{-\frac{1}{2}} \hat{\mathbf{A}} \hat{\mathbf{D}}^{-\frac{1}{2}} \ \mathbf{H}^{(v_1, ~l-1)}_{graph} \ \mathbf{W}^{(l-1)} \right),
\end{equation}
where $\hat{\mathbf{A}} = \mathbf{A} + \mathbf{I}_N$ is the adjacency matrix $\mathbf{A}$ of the input graph $G_i$ with added self-connections, and $\mathbf{I}_N$ is the identity matrix. $\hat{\mathbf{D}}$ is the degree matrix, $\mathbf{H}^{(v_1, l-1)}$ is node embedding matrix in the $(l-1)$-th layer of view $v_1$, $\mathbf{W}^{(l-1)}$ is a layer-specific trainable weight matrix, and $\sigma(\cdot)$ is a non-linear activation function \cite{2016_arXiv_GCN}. The calculation of view $v_2$ can be calculated in the same way.
After we obtain the final embedding $\mathbf{h}^{l}_i$ of node $i$ in tangent space, we map the final embedding from tangent space to hyperbolic space using the exponential map.

\subsubsection{Hyperbolic Hypergraph Aggregation}
Similar to hyperbolic graph aggregation, we first map the initial embedding $e^0_i$ in hyperbolic space to the tangent space using the logarithmic map, then
we employ HGCN as our fundamental hypergraph encoder to perform hypergraph convolution aggregation. The propagation rule in the $l$-th layer on the view $v_1$ can be expressed as:
\small
\begin{equation}
\begin{aligned}
    \mathbf{H}^{(v_1, ~l)}_{hyperg} = \sigma \big(&\mathbf{D}^{-1/2}_{hyperg} \ \mathbf{H}_{inc} \ \mathbf{W} \mathbf{B}^{-1} \ \mathbf{H}^T_{inc} \ \mathbf{D}^{-1/2}_{hyperg} \ \mathbf{H}^{(v_1, ~l-1)}_{hyperg} \ \mathbf{P}\big), 
\end{aligned}
\end{equation}
\normalsize
where $\mathbf{D}_{hyperg} \in \mathbb{R}^{N \times N}$ is the vertex degree matrix, $\mathbf{B} \in \mathbb{R}^{M \times M}$ is the hyperedge degree matrix, $\mathbf{W} \in \mathbb{R}^{M \times M}$ is the hyperedge weights matrix, $\mathbf{P} \in \mathbb{R}^{F^{(l-1)} \times F^{(l)}}$ is weight matirx between the $(l-1)$-th and $(l)$-th layer \cite{2021_PR_HGCN_HGAT_HyperGraph}. The calculation of view $v_2$ can be calculated in the same way. After we obtain the final embedding $\mathbf{h}^{l}_i$ of node $i$ in tangent space, we map the final embedding from tangent space to hyperbolic space using the exponential map.

\subsection{Multi-Level Contrast}
Following \cite{2023_WSDM_GOOD-D, 2023_ECMLPKDD_CVTGAD}, we design a contrastive strategy considering both node-level and graph-level contrast to train the model. Our proposed model comprises both graph- and hypergraph-channels, and their methods for computing multi-level contrast are similar. We elaborate on this as follows through graph-channel contrast.

\begin{algorithm}[htbp]
\caption{HC-GLAD}
\label{alg:algorithm}
\footnotesize
\LinesNumbered  
\SetKwInOut{Input}{Input}
\SetKwInOut{Output}{Output}
\SetKwInOut{Initialize}{Initialize}
\SetKwInOut{Parameter}{Parm}

\Input{Graph set:  $\mathcal{G} = \{G_1, G_2, ..., G_m \}$;}
\Output{The anomaly scores for each graph $Score_G$;}
\Initialize{(i) graph data augmentation: Obtain two augmented graphs (i.e., $v_1$ and $v_2$) using perturbation-free graph augmentation strategy \cite{2023_WSDM_GOOD-D, 2023_AAAI_FedStar}; (ii) hypergraph construction: Construct hypergraph by pre-designed gold motifs.}
\textbf{Training Phase} 
\For{$i = 1$ to $s\_epochs$}{
    Obtain initial hyperbolic node state $e^{0}$ by Eq.~(\ref{Equation: initial_hyperbolic_node_state}).\\
    Hyperbolic graph aggregation. \\
    Hyperbolic hypergraph Aggregation. \\
    Graph-channel:   \\
    {\hspace{0.5cm}(i) conduct node-level contrast by Eq.~(\ref{node_level_contrast});\\
    \hspace{0.5cm}(ii) conduct graph-level contrast by Eq.~(\ref{graph_level_contrast}).} \\
    Hypergraph-channel:  \\ 
    {\hspace{0.5cm}(i) conduct node-level contrast by Eq.~(\ref{node_level_contrast}); \\ 
    \hspace{0.5cm}(ii) conduct graph-level contrast by Eq.~(\ref{graph_level_contrast}).} \\
    Calculate graph-channel loss by Eq.~(\ref{graph-channel_loss}). \\
    Calculate hypergraph-channel loss similarly to the way to calculate graph-channel loss. \\
    Calculate the total loss by Eq.~(\ref{Equation: Loss_total}).
}
\textbf{Inference Phase}
\For{$G_{i}$ in Graph set $G$}{
    Calculate anomaly scores by Eq.~(\ref{Equation: Anomaly Score}).
}
\end{algorithm}

\textbf{Node-level Contrast.} For an input graph $G_i$, we first map node embedding into node-level contrast space with MLP-based projection head, and then we construct node-level contrastive loss to maximize the agreement between the embeddings belonging to different views on the node level:

\small
\begin{equation}
\begin{aligned}
    \mathcal{L}_{node} = & \frac{1}{|\mathcal{B}|} \sum_{G_j \in \mathcal{B}} \frac{1}{2|\mathcal{V}_{G_j}|}
    \sum_{v_i \in \mathcal{V}_{G_j}} 
    \Big[ 
    \emph{l}\left(\mathbf{h}^{(v_1)}_{i}, \mathbf{h}^{(v_2)}_{i}\right) + \emph{l}\left(\mathbf{h}^{(v_2)}_{i}, \mathbf{h}^{(v_1)}_{i}\right)\Big],    
\end{aligned}
\label{node_level_contrast}
\end{equation}
\normalsize
\begin{equation}
\begin{split}
    \emph{l}\left(\mathbf{h}^{(v_1)}_{i}, \mathbf{h}^{(v_2)}_{i}\right) = 
    -\log \frac{e^{\left(-H_{Dist}\left(\mathbf{h}^{(v_1)}_{i}, \mathbf{h}^{(v_2)}_{i}\right) / {\tau}\right)}}{\sum_{v_{k} \in \mathcal{V}_{G_j} \backslash v_i} e^{\left(-H_{Dist}\left(\mathbf{h}^{(v_1)}_{i}, \mathbf{h}^{(v_2)}_{k}\right) / {\tau}\right)}}.
\end{split}
\label{contrast_calculate}
\end{equation}

In Eq. (\ref{node_level_contrast}), $\mathcal{B}$ is the training/testing batch and $\mathcal{V}_{G_j}$ is the node set of graph $G_j$. The calculation of $\emph{l}\left(\mathbf{h}^{(v_{2})}_{i}, \mathbf{h}^{(v_{1})}_{i}\right)$ and $\emph{l}\left(\mathbf{h}^{(v_{1})}_{i}, \mathbf{h}^{(v_{2})}_{i}\right)$ is the same, and we briefly show the calculation of $\emph{l}\left(\mathbf{h}^{(v_{1})}_{i}, \mathbf{h}^{(v_{2})}_{i}\right)$ in Eq. (\ref{contrast_calculate}). In Eq. (\ref{contrast_calculate}), the $H_{Dist}\left(.,.\right)$ is the function to measure the hyperbolic distance between different views. In this work, we compute the distance as Eq.~(\ref{equ: Lorentzian_distance}) indicates.

\begin{table*}[!ht]
\caption{The hyperbolicity $\delta$ and average hyperbolicity $\delta_{avg}$ of datasets.}
\vspace{-0.5em}
\label{hyperbolicity_result}
\centering
\fontsize{7.5}{8}\selectfont   
    \renewcommand{\arraystretch}{1.1}
    \begin{tabular}{c|ccccccccccccc}
    \toprule[1.5pt]

    \textbf{Dataset}    & \textbf{PROTEINS\_full} & \textbf{ENZYMES} & \textbf{AIDS} & \textbf{DHFR} & \textbf{BZR} & \textbf{COX2} & \textbf{DD} & \textbf{REDDIT-B} & \textbf{HSE} & \textbf{MMP} & \textbf{p53} & \textbf{PPAR-gamma}  & \textbf{IMDB-B} \\ 
    \midrule[1pt]
    \bm{$\delta$} & 1.09        & 1.15     & 0.74        & 1.01     &  1.11            & 1.00          & 3.74          & 0.97     & 0.76      & 0.77  & 0.78    & 0.77    & 0.24 \\
    \midrule
    \bm{$\delta_{avg}$}       & 0.14        & 0.15     & 0.15    &  0.12     & 0.18        & 0.09      & 0.64       & 0.05       & 0.12    & 0.12  & 0.12    & 0.12     & 0.02 \\ 
    \bottomrule[1.5pt]
    \end{tabular}
\end{table*}

\begin{table*}[]
\caption{The performance comparison in terms of AUC (in percent, mean value ± standard deviation). The best performance is highlighted in \textbf{\textcolor{darkblue}{bold}}, and the second-best performance is \textcolor{lightblue}{{\ul underlined}}. \dag: we report the result from \cite{2023_WSDM_GOOD-D}.}
\vspace{-0.5em}
\label{overall_performance}
\centering
\scalebox{0.78}{
    \renewcommand{\arraystretch}{1.44}
    \begin{tabular}{l|ccccccccc>{\columncolor{gray!20}}c}
    \toprule[2pt]
    \textbf{Method}                 & \textbf{PK-OCSVM\dag} & \textbf{PK-iF\dag} & \textbf{WL-OCSVM\dag} & \textbf{WL-iF\dag} & \textbf{InfoGraph-iF\dag} & \textbf{GraphCL-iF\dag} & \textbf{OCGIN\dag}      & \textbf{GLocalKD\dag}   & \textbf{GOOD-D\dag}     & \textbf{HC-GLAD}     \\ \midrule[1.5pt]
    \textbf{PROTEINS-full} & 50.49±4.92        & 60.70±2.55     & 51.35±4.35        & 61.36±2.54     & 57.47±3.03            & 60.18±2.53          & 70.89±2.44          & \textcolor{lightblue}{\ul{77.30±5.15}} & 71.97±3.86          & \textbf{\textcolor{darkblue}{77.51±2.58}}    \\
    \textbf{ENZYMES}       & 53.67±2.66        & 51.30±2.01     & 55.24±2.66        & 51.60±3.81     & 53.80±4.50            & 53.60±4.88          & 58.75±5.98          & 61.39±8.81          & \textcolor{lightblue}{\ul{63.90±3.69}}    & \textbf{\textcolor{darkblue}{65.39±6.23}} \\
    \textbf{AIDS}          & 50.79±4.30        & 51.84±2.87     & 50.12±3.43        & 61.13±0.71     & 70.19±5.03            & 79.72±3.98          & 78.16±3.05          & 93.27±4.19          & \textcolor{lightblue}{\ul{97.28±0.69}}    & \textbf{\textcolor{darkblue}{99.51±0.38}} \\
    \textbf{DHFR}          & 47.91±3.76        & 52.11±3.96     & 50.24±3.13        & 50.29±2.77     & 52.68±3.21            & 51.10±2.35          & 49.23±3.05          & 56.71±3.57          & \textbf{\textcolor{darkblue}{62.67±3.11}}    & \textcolor{lightblue}{\ul{61.43±4.27}} \\
    \textbf{BZR}           & 46.85±5.31        & 55.32±6.18     & 50.56±5.87        & 52.46±3.30     & 63.31±8.52            & 60.24±5.37          & 65.91±1.47          & 69.42±7.78          & \textcolor{lightblue}{\ul{75.16±5.15}}    & \textbf{\textcolor{darkblue}{75.75±9.11}} \\
    \textbf{COX2}          & 50.27±7.91        & 50.05±2.06     & 49.86±7.43        & 50.27±0.34     & 53.36±8.86            & 52.01±3.17          & 53.58±5.05          & 59.37±12.67         & \textbf{\textcolor{darkblue}{62.65±8.14}}    & \textcolor{lightblue}{\ul{59.98±7.44}} \\
    \textbf{DD}            & 48.30±3.98        & 71.32±2.41     & 47.99±4.09        & 70.31±1.09     & 55.80±1.77            & 59.32±3.92          & 72.27±1.83          & \textbf{\textcolor{darkblue} {80.12±5.24}} & 73.25±3.19          & \textcolor{lightblue}{\ul{77.66±1.73}}    \\
    \textbf{REDDIT-B}      & 45.68±2.24        & 46.72±3.42     & 49.31±2.33        & 48.26±0.32     & 68.50±5.56            & 71.80±4.38          & 75.93±8.65          & 77.85±2.62          & \textbf{\textcolor{darkblue} {88.67±1.24}} & \textcolor{lightblue}{\ul{79.09±2.52}}    \\
    \textbf{HSE}           & 57.02±8.42        & 56.87±10.51    & 62.72±10.13       & 53.02±5.12     & 53.56±3.98            & 51.18±2.71          & \textcolor{lightblue}{\ul{64.84±4.70}}          & 59.48±1.44          & \textbf{\textcolor{darkblue}{69.65±2.14}}    & 64.05±4.75 \\
    \textbf{MMP}           & 46.65±6.31        & 50.06±3.73     & 55.24±3.26        & 52.68±3.34     & 54.59±2.01            & 54.54±1.86          & \textbf{\textcolor{darkblue}{71.23±0.16}}    & 67.84±0.59          & 70.51±1.56          & \textcolor{lightblue}{\ul{70.96±4.45}}    \\
    \textbf{p53}           & 46.74±4.88        & 50.69±2.02     & 54.59±4.46        & 50.85±2.16     & 52.66±1.95            & 53.29±2.32          & 58.50±0.37          & \textcolor{lightblue}{\ul{64.20±0.81}}    & 62.99±1.55          & \textbf{\textcolor{darkblue}{66.01±1.77}} \\
    \textbf{PPAR-gamma}    & 53.94±6.94        & 45.51±2.58     & 57.91±6.13        & 49.60±0.22     & 51.40±2.53            & 50.30±1.56          & \textbf{\textcolor{darkblue}{71.19±4.28}} & 64.59±0.67          & 67.34±1.71          & \textcolor{lightblue}{\ul{69.51±5.04}}    \\ 
    \textbf{IMDB-B} & 50.75±3.10        & 50.80±3.17     & 54.08±5.19        & 54.08±5.19     & 56.50±3.58            & 56.50±4.90          & 60.19±8.90          & 52.09±3.41              & \textbf{\textcolor{darkblue}{65.88±0.75}}     & \textcolor{lightblue}{\ul{60.92±3.39}}\\
    \midrule
    \textbf{Avg.Rank}               & 8.77           & 7.85           & 7.15           & 7.77        & 6.15               & 6.54             & 3.77             & 3.31             & \textcolor{lightblue}{\ul{2.00}}                & \textbf{\textcolor{darkblue}{1.69}}               \\
    \bottomrule[2pt]
    \end{tabular}
}
\end{table*}

\textbf{Graph-level Contrast.} To obtain graph embedding $\mathbf{h}_{G_i}$ of graph $G_i$, we employ the pooling operation simply on embeddings of nodes in graph $G_i$. We first map graph embedding into graph-level contrast space with an MLP-based projection head. Similar to the node-level loss $\mathcal{L}_{node}$, we then construct a graph-level loss for mutual agreement maximization on the graph level:
\begin{equation}
    \mathcal{L}_{graph} = \frac{1}{2|\mathcal{B}|} \sum_{G_i \in \mathcal{B}} \left[\emph{l}\left(\mathbf{h}^{(v_1)}_{G_i}, \mathbf{h}^{(v_2)}_{G_i}\right) + \emph{l}\left(\mathbf{h}^{(v_2)}_{G_i}, \mathbf{h}^{(v_1)}_{G_i}\right) \right],
\label{graph_level_contrast}
\end{equation}
\begin{equation}
    \emph{l}\left(\mathbf{h}^{(v_1)}_{G_i}, \mathbf{h}^{(v_2)}_{G_i}\right) = -log \frac{e^{\left(-H_{Dist}\left(\mathbf{h}^{(v_1)}_{G_i}, \mathbf{h}^{(v_2)}_{G_i}\right) /  {\tau}\right)}}{\sum_{G_j \in \mathcal{B} \backslash G_i} e^{\left(-H_{Dist}\left(\mathbf{h}^{(v_1)}_{G_i}, \mathbf{h}^{(v_2)}_{G_j}\right) / \tau\right)}},
\end{equation}
where notations are similar to node-level loss, and $\emph{l}\left(\mathbf{h}^{(v_2)}_{G_i}, \mathbf{h}^{(v_1)}_{G_i}\right)$ is calculated in the same way as $\emph{l}\left(\mathbf{h}^{(v_1)}_{G_i}, \mathbf{h}^{(v_2)}_{G_i}\right)$. The training loss function on the graph channel is:
\begin{equation}
    \mathcal{L}_{graph-channel} = \xi_1~ \mathcal{L}_{node} + \xi_2~ \mathcal{L}_{graph},
\label{graph-channel_loss}
\end{equation}
where $\xi_1$ and $\xi_2$ are trade-off parameters, and we set $\xi_1 = 1$ and $\xi_2 = 1$ on experiments of this work for simplicity.
The training loss function on the hypergraph channel is calculated in the same way as the one on the graph channel. Therefore, in the training phase, we employ the loss function as:
\begin{equation}
    \mathcal{L}_{total} = \lambda_1~ \mathcal{L}_{graph-channel} + \lambda_2~ \mathcal{L}_{hypergraph-channel}.
\label{Equation: Loss_total}
\end{equation}
\subsection{Anomaly Scoring}
In the inference phase, we calculate the anomaly score from both graph-channel and hypergraph-channel, where a higher score indicates a greater anomaly. For simplicity and efficiency, we directly employ the $\mathcal{L}_{total}$ (Eq. (\ref{Equation: Loss_total})) as the final anomaly score for an input graph $G_i$ as:
\begin{equation}
    score_{G_i} = \mathcal{L}_{total} .
\label{Equation: Anomaly Score}
\end{equation}

\section{Experiments}

\begin{figure*}[]
    \centering
    \vspace{-1.5em}
    \subfigure{\includegraphics[width=0.19\textwidth]{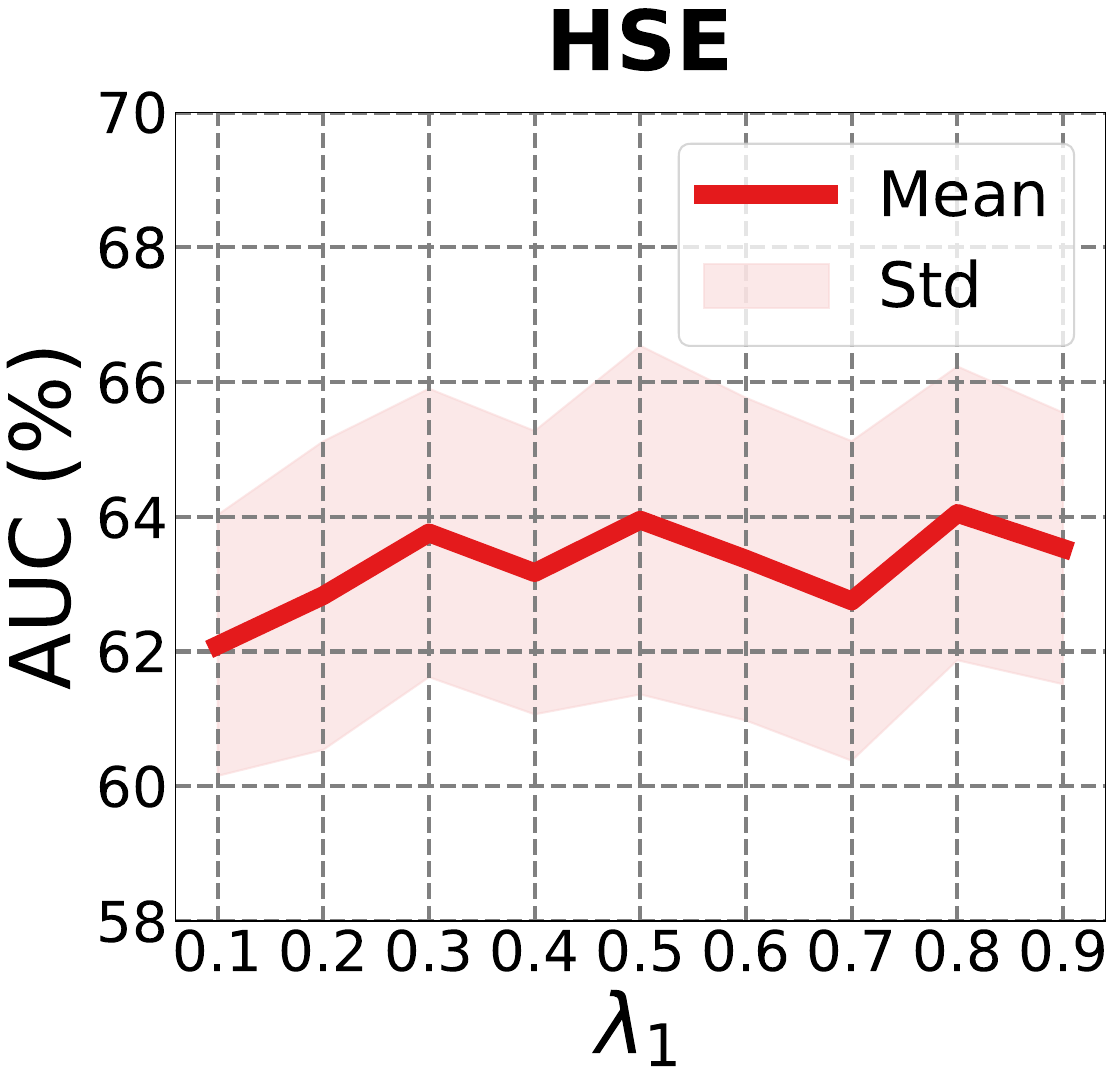}}
    \hspace{-0.1cm}
    \subfigure{\includegraphics[width=0.19\textwidth]{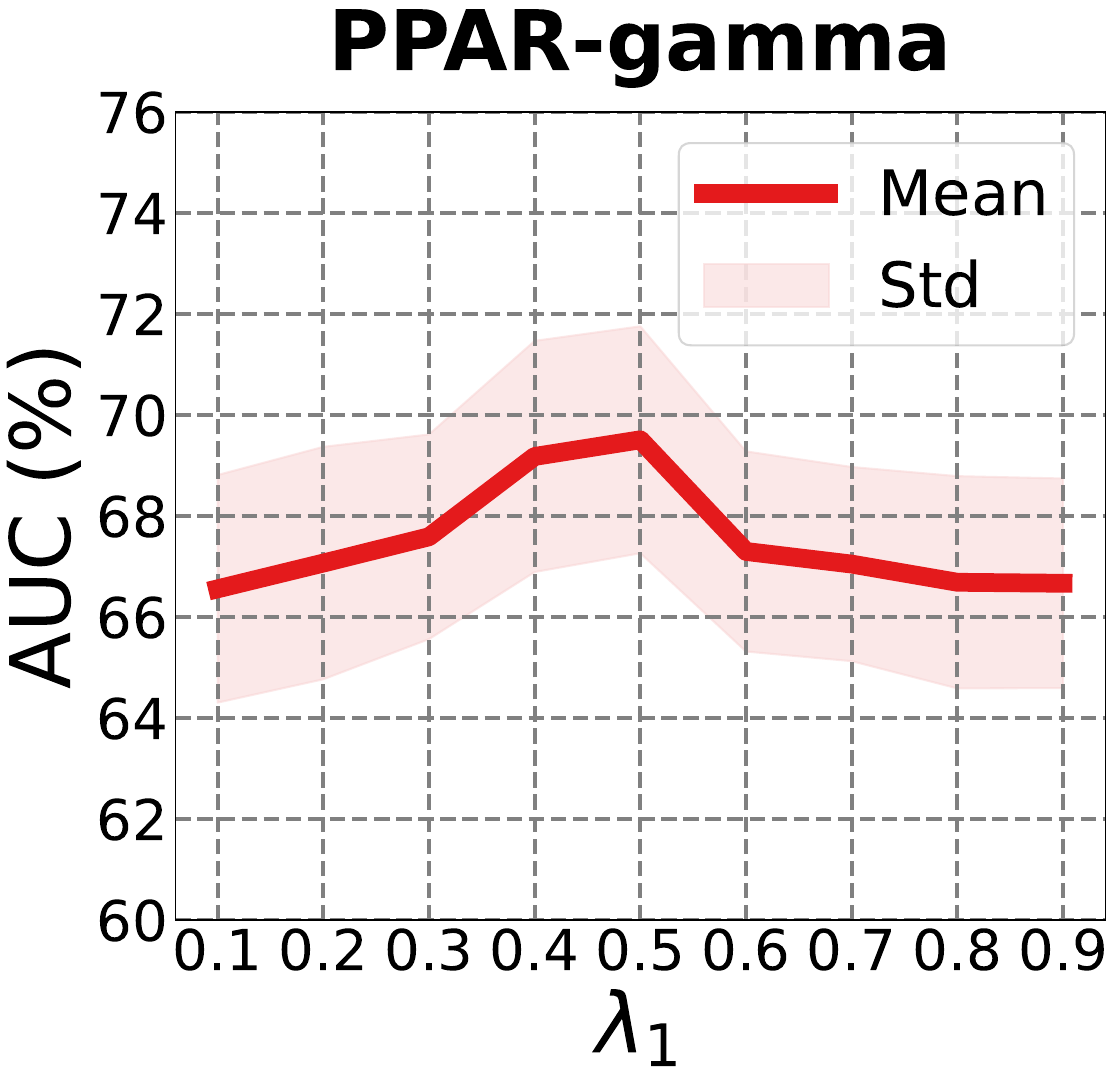}}
    \hspace{-0.1cm}
    \subfigure{\includegraphics[width=0.202\textwidth]{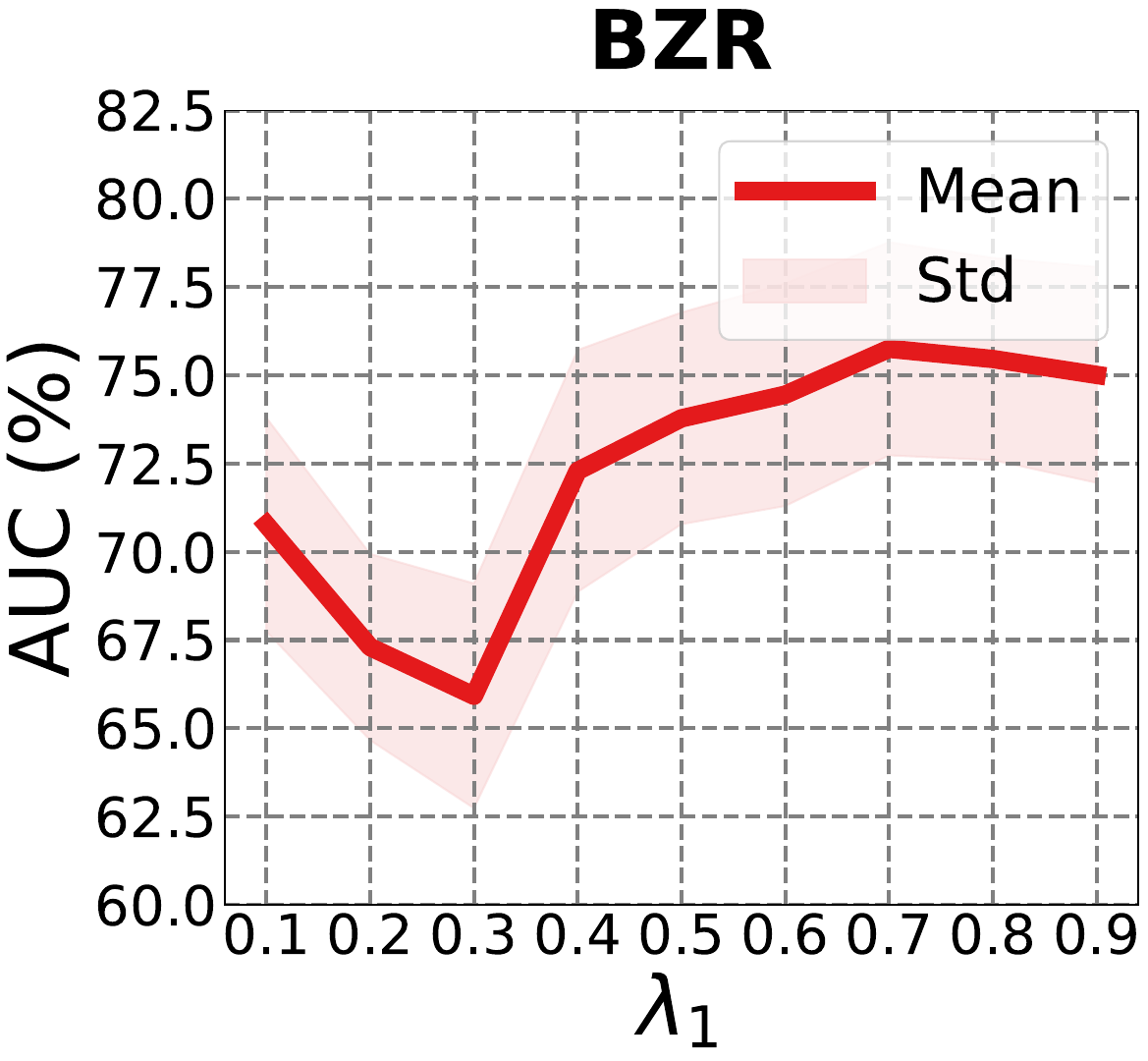}}
    \hspace{-0.1cm}
    \subfigure{\includegraphics[width=0.211\textwidth]{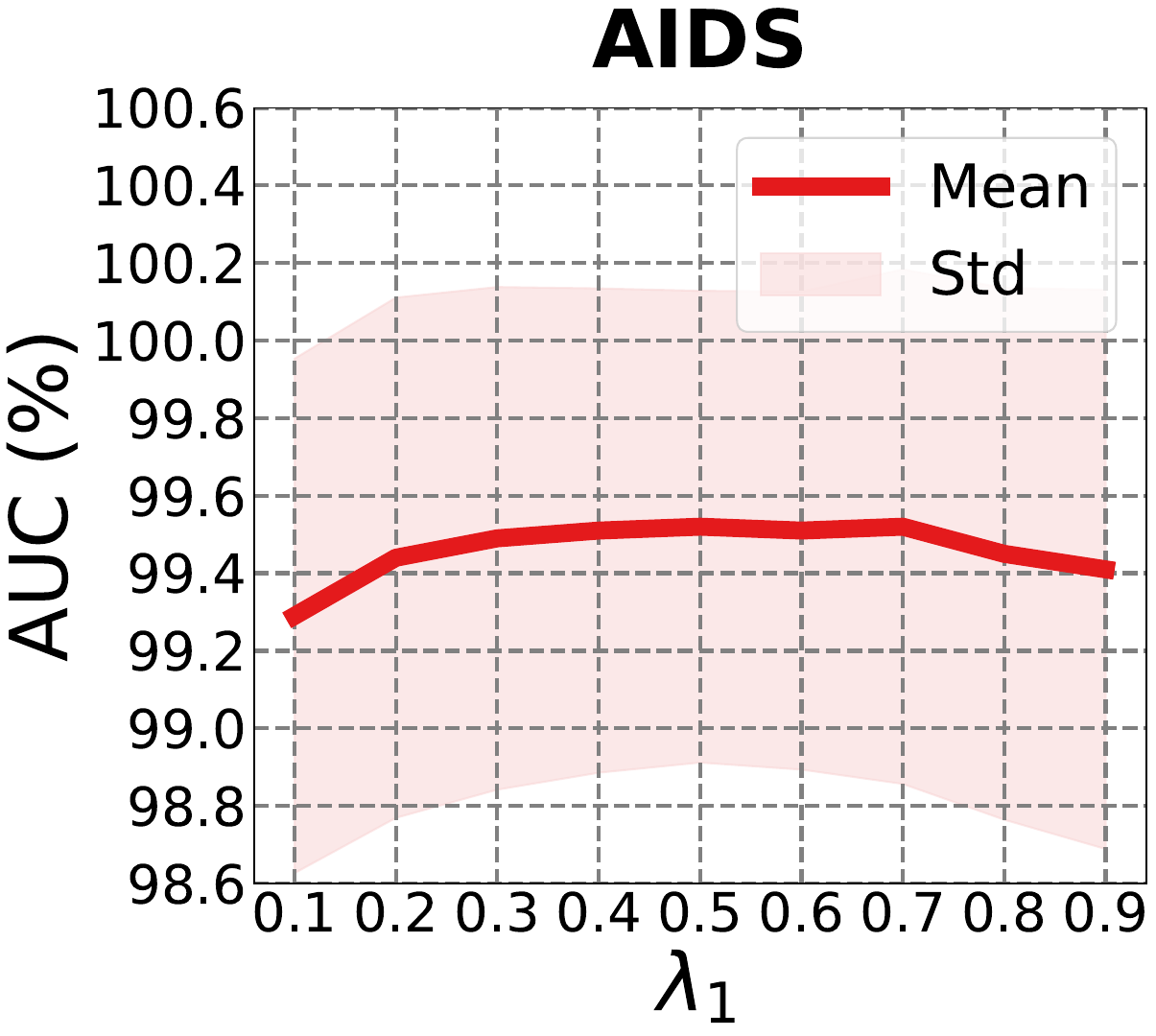}}
    \hspace{-0.1cm}
    \subfigure{\includegraphics[width=0.195\textwidth]{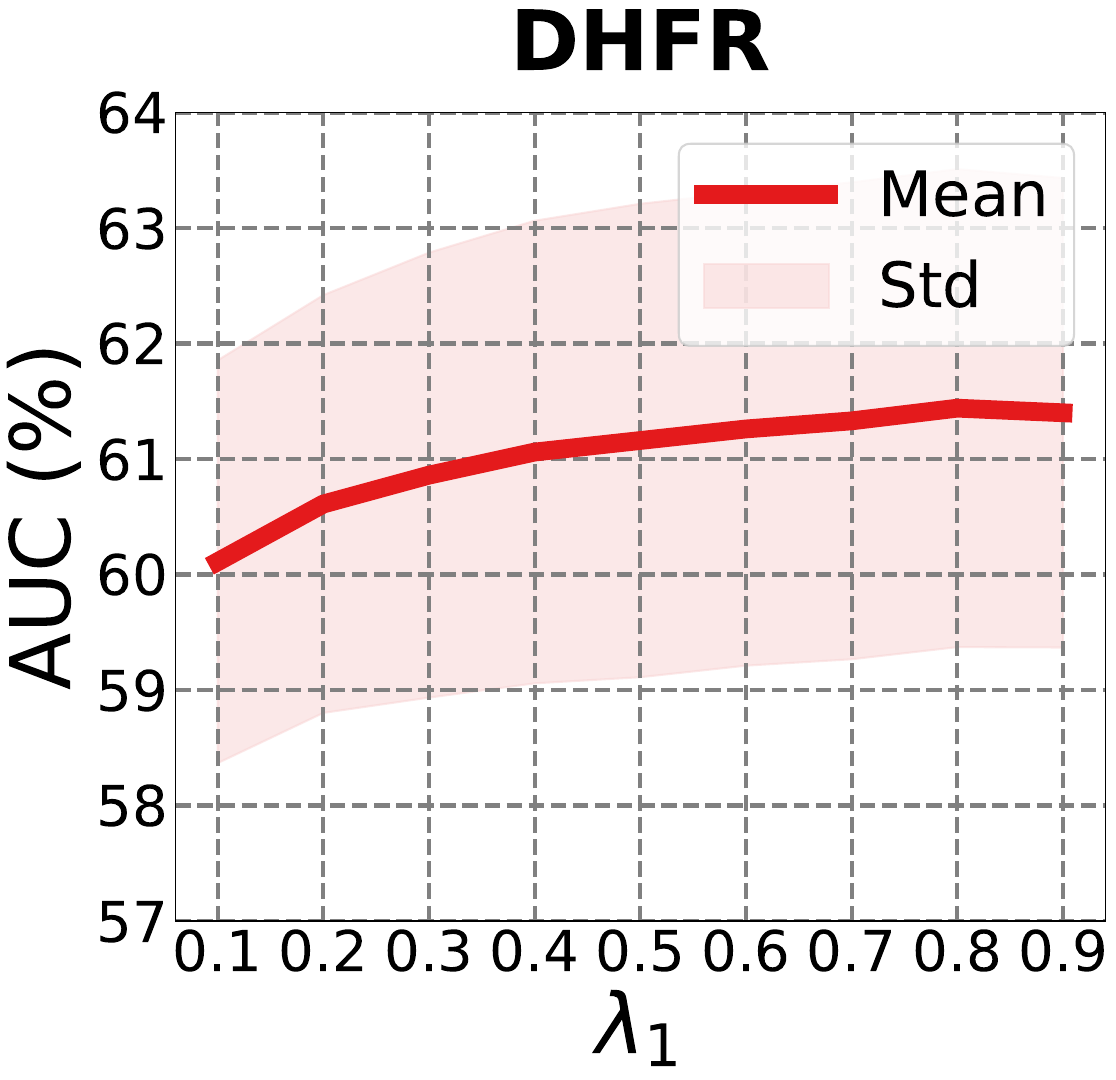}}
\vspace{-1.5em}
\caption{Hyper-parameter analysis (trade-off parameter $\lambda_1$) on representative datasets.}
\label{fig: Hyper-parameter-lambda_1}
\end{figure*}

\subsection{Experimental Setup}
\textbf{Datasets.} We conduct experiments on 13 open-source datasets from TUDataset \cite{2020_arXiv_TuDataset}, which involves small molecules, bioinformatics, and social networks. Appendix~\ref{appendix:datasets_details} provides more details of the datasets. We follow the settings in \cite{2022_WSDM_GLocalKD, 2023_WSDM_GOOD-D} to define anomaly, while the rest are viewed as normal data (i.e., normal graphs). Similar to \cite{2022_WSDM_GLocalKD, 2023_WSDM_GOOD-D, 2021_BigData_OCGIN}, only normal data are utilized during the training phase. 

To measure the hyperbolic nature in the datasets, we introduce the hyperbolicity $\delta$ proposed by Gromov \cite{1987_Hyperbolic_Groups}. In general, the hyperbolicity $\delta$ quantifies the tree-likeness of a graph. The lower the value of $\delta$, the more tree-like the structure, suitable to embed in hyperbolic space \cite{2019_ICLR_GLOVE}. When $\delta$ = 0, the graph can be considered a tree \cite{2014_define_avg_hyperbolicity,2023_TKDE_HTCN}. For more accuracy, we also report the average hyperbolicity $\delta_{avg}$, which is robust to the addition or removal of an edge from the graph \cite{2021_WWW_LGCN}. Given that the time complexity for calculating $\delta$ and $\delta_{avg}$ is O($n^{4}$), we employ a random sampling method to approximate the calculations \cite{2019_NeurIPS_HGCN_Hyperbolic,2021_BigData_HAT_Hyperbolic,2021_WWW_LGCN}. The results are illustrated in Table \ref{hyperbolicity_result}. Appendix~\ref{appendix:hyperbolic_definition} explains the definition of hyperbolicity and shows the hyperbolicity distribution of the graph for some datasets in detail.

\textbf{Baselines.} We select 9 representative baselines from the non-end-to-end and end-to-end methods to compare with our proposed model. And for the non-end-to-end methods, we mainly select two categories: (i) kernel + detector. (ii) GCL model + detector. 

\begin{figure}[]
    \centering
    \subfigure{\includegraphics[width=0.195\textwidth]{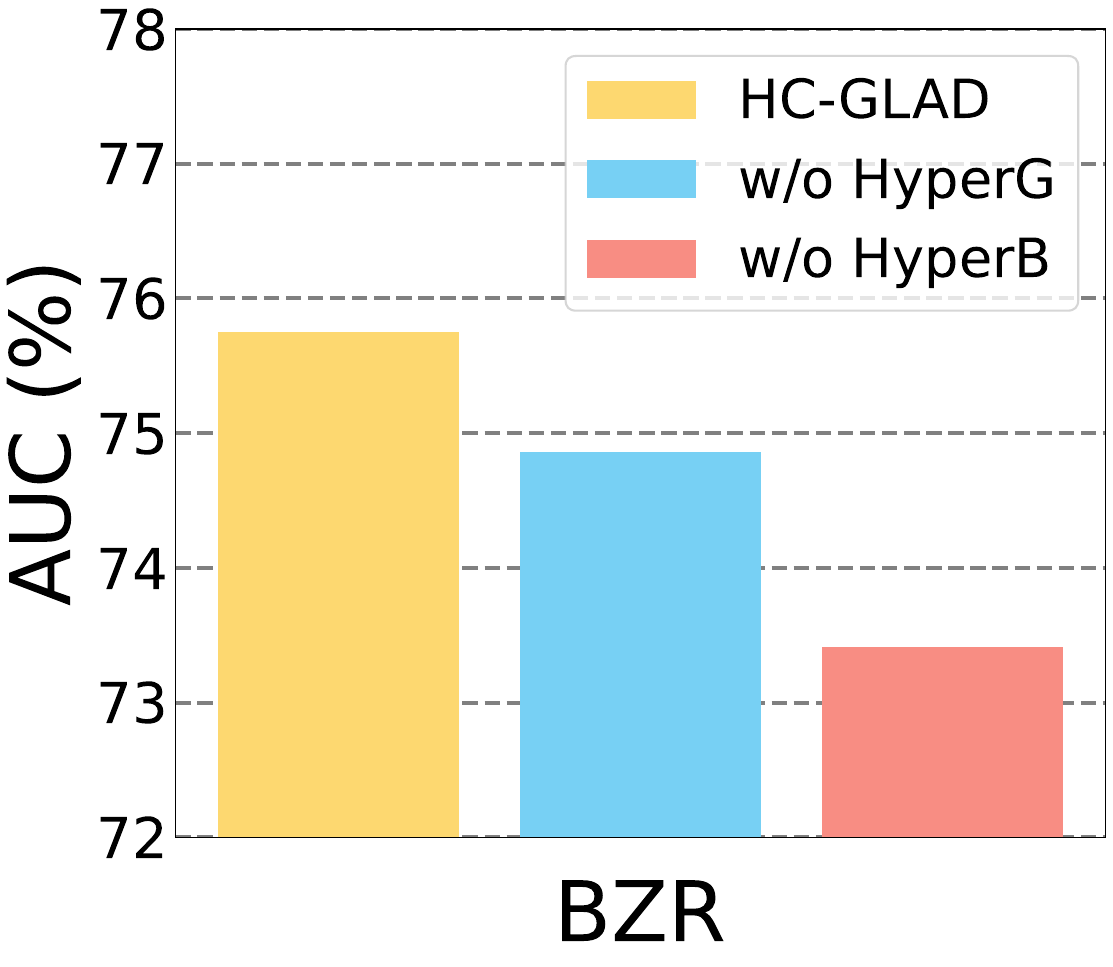}}
    \hspace{2mm}
    \subfigure{\includegraphics[width=0.195\textwidth]{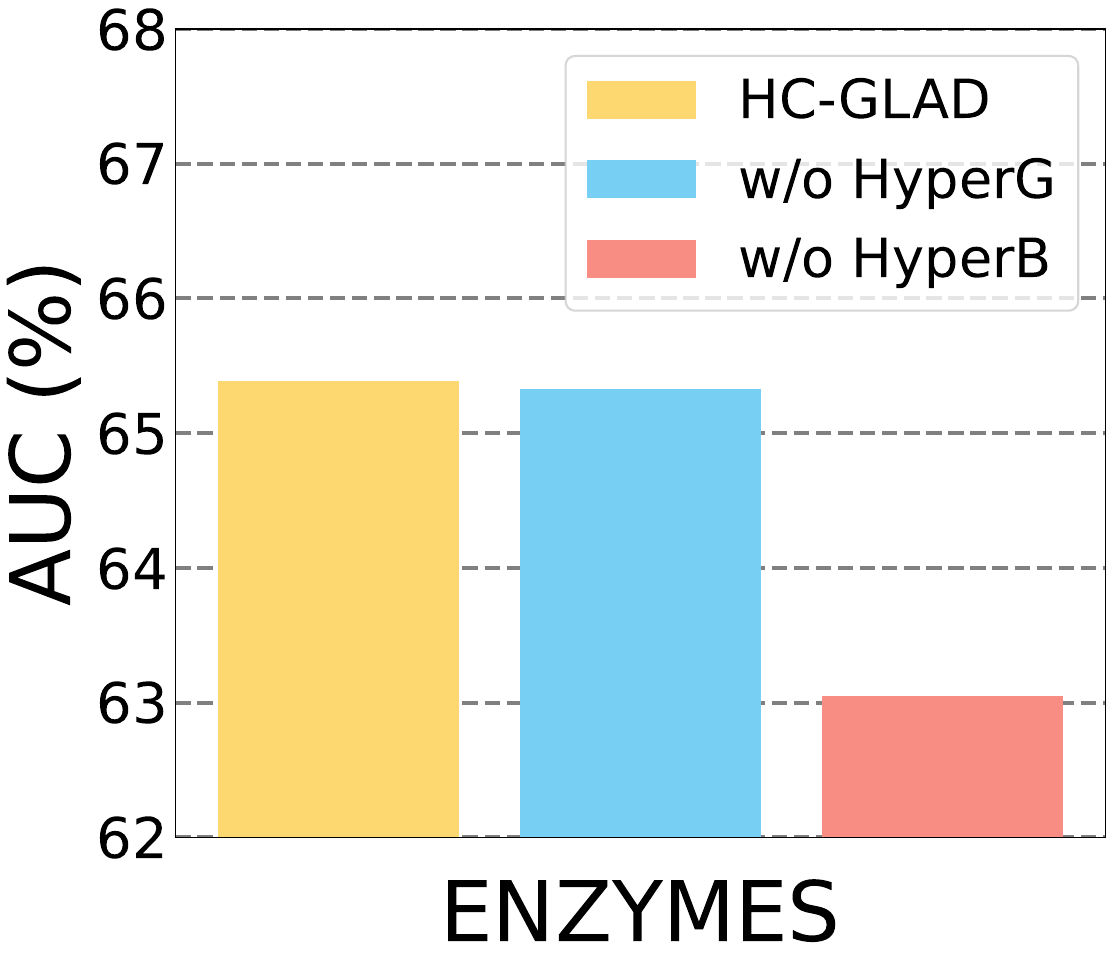}}
    \hspace{2mm}
    \subfigure{\includegraphics[width=0.195\textwidth]{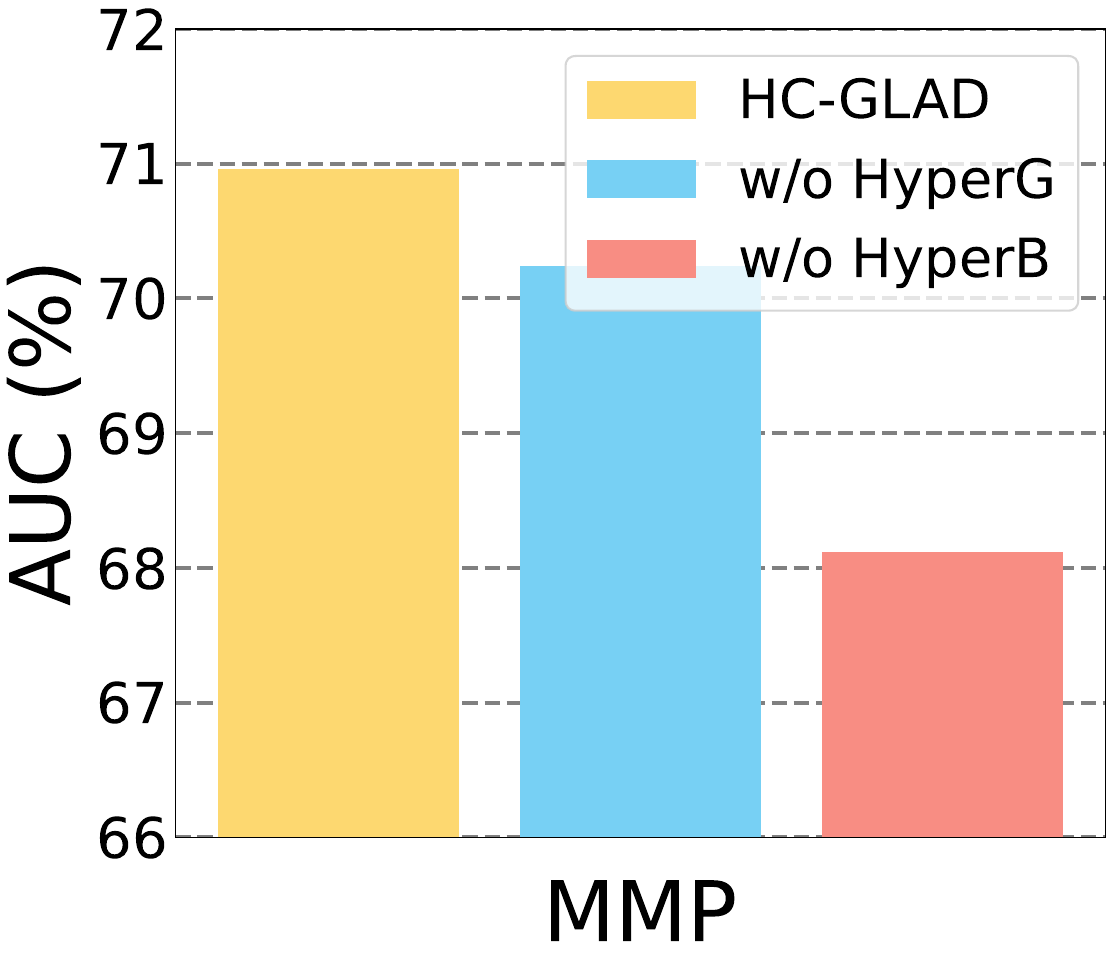}}
    \hspace{2mm}
    \subfigure{\includegraphics[width=0.195\textwidth]{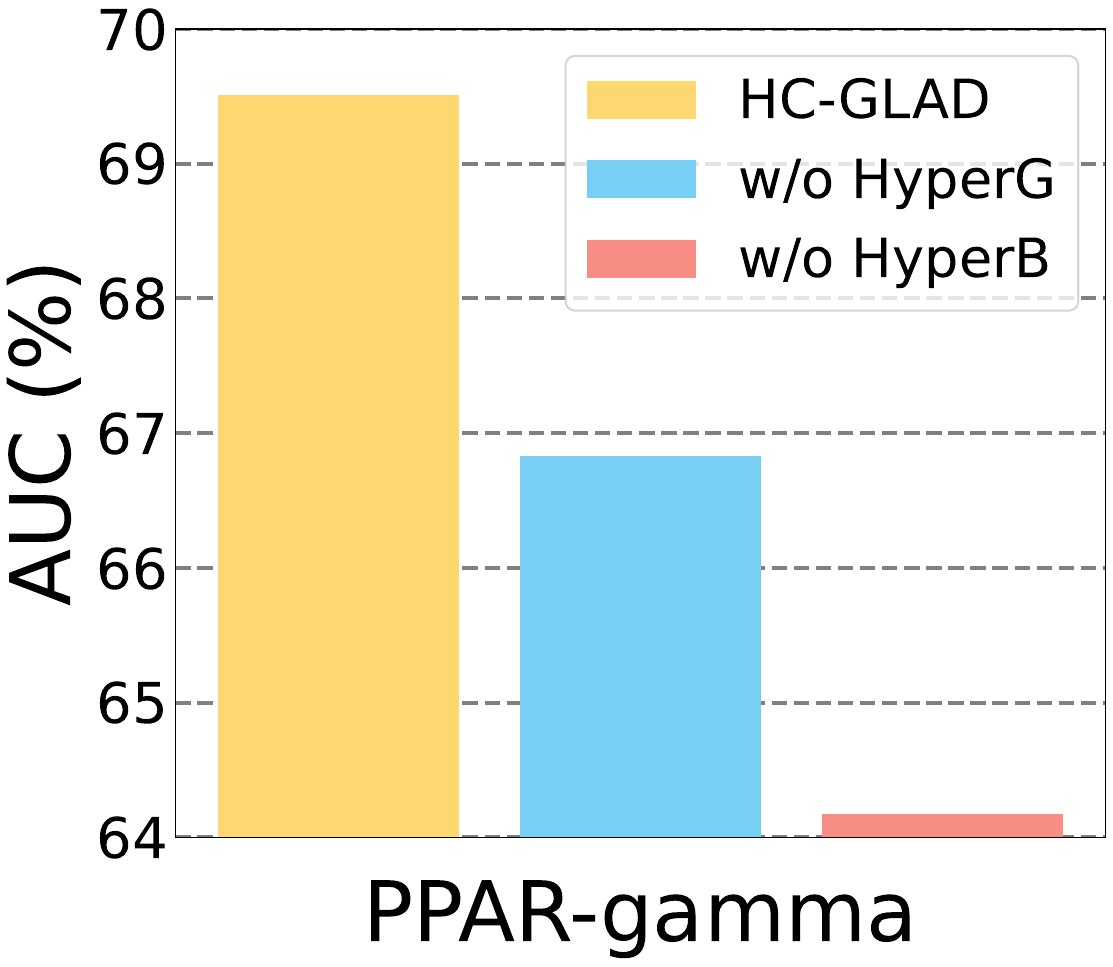}}
\vspace{-1em}
\caption{Ablation study on representative datasets.}
\label{fig: Ablation-study}
\vspace{-1em}
\end{figure}

\begin{itemize}[leftmargin=*,noitemsep,topsep=1.5pt]
    \item \textbf{Graph kernel + detector}. We adopt Weisfeiler-Lehman kernel (WL in short) \cite{2011_JMLR_WL_graph_Kernel} and propagation kernel (PK in short) \cite{2016_ML_PK_graph_Kernel} to first obtain representations, and then we take one-class SVM (OCSVM in short) \cite{2001_JMLR_OCSVM} and isolation forest (iF in short) \cite{2008_ICDM_iF_graph_Kernel} to detect anomaly. After arranging and combining the above kernels and detectors, there are four baselines available: PK-OCSVM, PK-iF, WL-OCSVM, and WL-iF.
    \item \textbf{GCL model + detector}. Considering that we used the paradigm of graph contrastive learning, we select two classic graph-level contrastive learning models (i.e., InfoGraph \cite{2020_ICLR_InfoGraph} and GraphCL~\cite{2020_NeurIPS_GraphCL}) to first obtain representations, and then we take iF as detector to detect anomaly (i.e., InfoGraph-iF, GraphCL-iF).
    \item \textbf{End-to-end method}. We select 3 classical models: OCGIN \cite{2021_BigData_OCGIN}, GLocalKD \cite{2022_WSDM_GLocalKD} and GOOD-D \cite{2023_WSDM_GOOD-D}. 
\end{itemize}

\textbf{Metrics and Implementations.} Following \cite{2022_WSDM_GLocalKD, 2022_ScientificReports_GLADC, 2023_WSDM_GOOD-D, 2023_ECMLPKDD_CVTGAD}, we adopt popular graph-level anomaly detection metric AUC (i.e., the area under the receiver operating characteristic) to evaluate methods. A higher AUC value corresponds to better anomaly detection performance. We use the Riemannian SGD with weight decay to learn the parameters of the network \cite{2022_KDD_HICF, 2013_TAC_Riemannian_SGD}. In practice, we implement HC-GLAD with PyTorch \cite{2019_NeurIPS_PyTorch_Library}.

\subsection{Overall Performance}
The AUC results of HC-GLAD, along with nine other baseline methods, are summarized in Table~\ref{overall_performance}. As depicted in Table~\ref{overall_performance}, HC-GLAD outperforms other methods by securing first place on 5 datasets and second place on 7 datasets, while maintaining a competitive performance on the remaining dataset. Furthermore, HC-GLAD achieves the best average rank among all methods across the 13 datasets. Our observations indicate that graph kernel-based methods exhibit the poorest performance among baselines. This underperformance is attributed to their limited ability to identify regular patterns and essential graph information, rendering them less effective with complex datasets. GCL-based methods show a moderate level of performance, highlighting the competitive potential of graph contrastive learning for UGAD tasks. 
In conclusion, the competitive performance of our proposed model underscores the effectiveness of incorporating node group information, as well as integrating hypergraph learning and hyperbolic geometry into graph-level anomaly detection. These findings also validate that HC-GLAD possesses inherent capabilities to capture fundamental characteristics of normal graphs, consequently delivering superior anomaly detection performance.

\begin{table}[]
\caption{The AUC (\%) performance comparison of different motifs for hypergraph construction.}
\vspace{-1em}
\label{motif_variant}
\centering
\scalebox{0.735}{
    \renewcommand{\arraystretch}{1.5}
    \begin{tabular}{c|cccccc}
    \toprule[1.5pt]
    \textbf{Dataset}   & \textbf{ENZYMES}    & \textbf{COX2}   & \textbf{PPAR-gamma}   & \textbf{p53}  & \textbf{BZR}  & \textbf{MMP} \\ 
    \midrule[1pt]
    \textbf{Triangle Variant}    & 64.23  & 57.66    & 68.71     & 65.49     & 75.51   & 70.27  \\     
    \textbf{Triangle (Ours)}   & \textbf{65.39}   & \textbf{59.98}     & \textbf{69.51}     & \textbf{66.01}     & \textbf{75.75}   & \textbf{70.96} \\
    \bottomrule[1.5pt]
    \end{tabular}
}
\vspace{-1em}
\end{table}

\begin{figure*}[t]
    \centering
    \vspace{-0.2em}
    \subfigure[$v_1$ of graph-channel]{\includegraphics[width=0.19\textwidth]{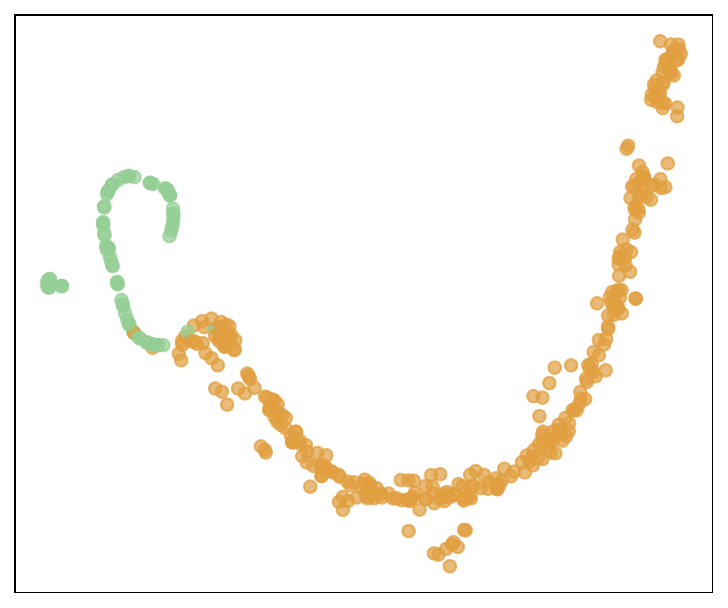}}
    \hspace{0.1mm}
    \subfigure[$v_2$ of graph-channel]{\includegraphics[width=0.19\textwidth]{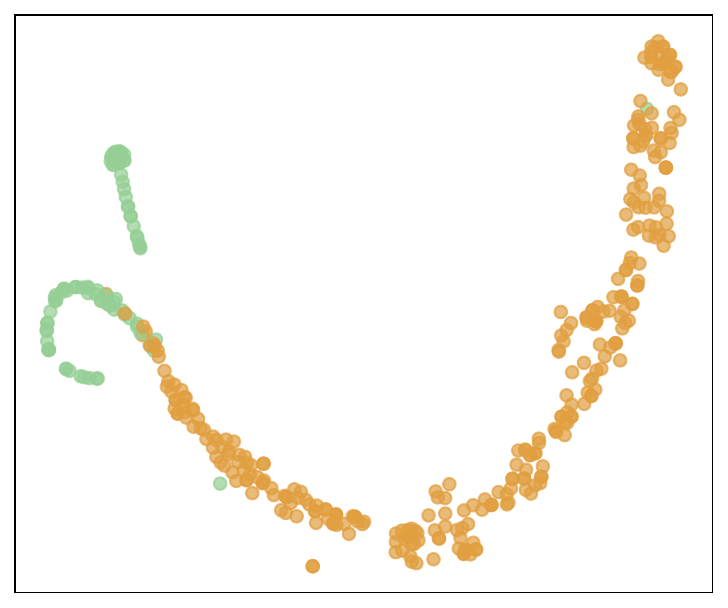}} 
    \hspace{0.1mm}
    \subfigure[$v_1$ of hypergraph-channel]{\includegraphics[width=0.19\textwidth]{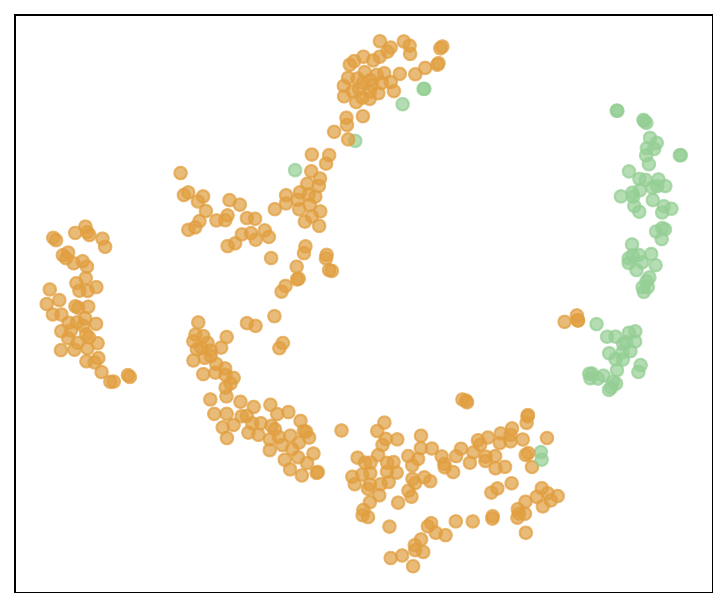}}
    \hspace{0.1mm}
    \subfigure[$v_2$ of hypergraph-channel]{\includegraphics[width=0.19\textwidth]{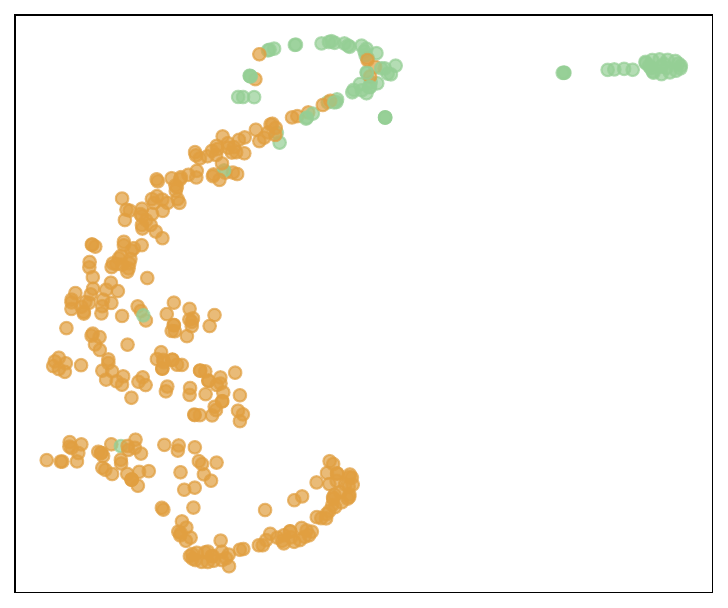}}
    \hspace{0.1mm}
    \subfigure[anomaly score]{\includegraphics[width=0.19\textwidth]{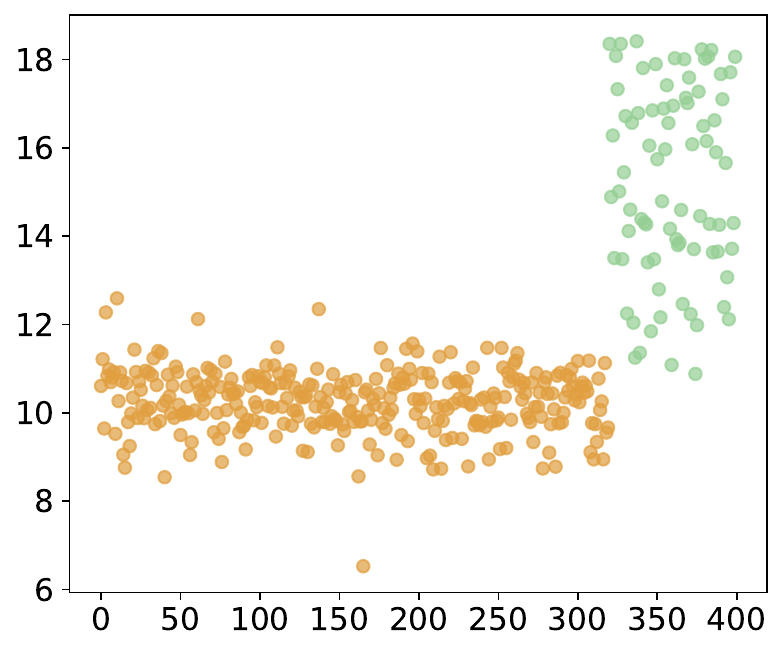}}
\vspace{-1em}
\caption{Visualization on AIDS dataset for different view $v_1$ and $v_2$. (\textcolor{visualization_yellow}{$\bullet$} denotes normal graph, \textcolor{visualization_green}{$\bullet$} denotes anomalous graph.)}
\label{fig: Visualization}
\end{figure*}

\begin{figure}
    \centering
    \subfigure[Different number of encoder layers.]{\includegraphics[width=0.23\textwidth]{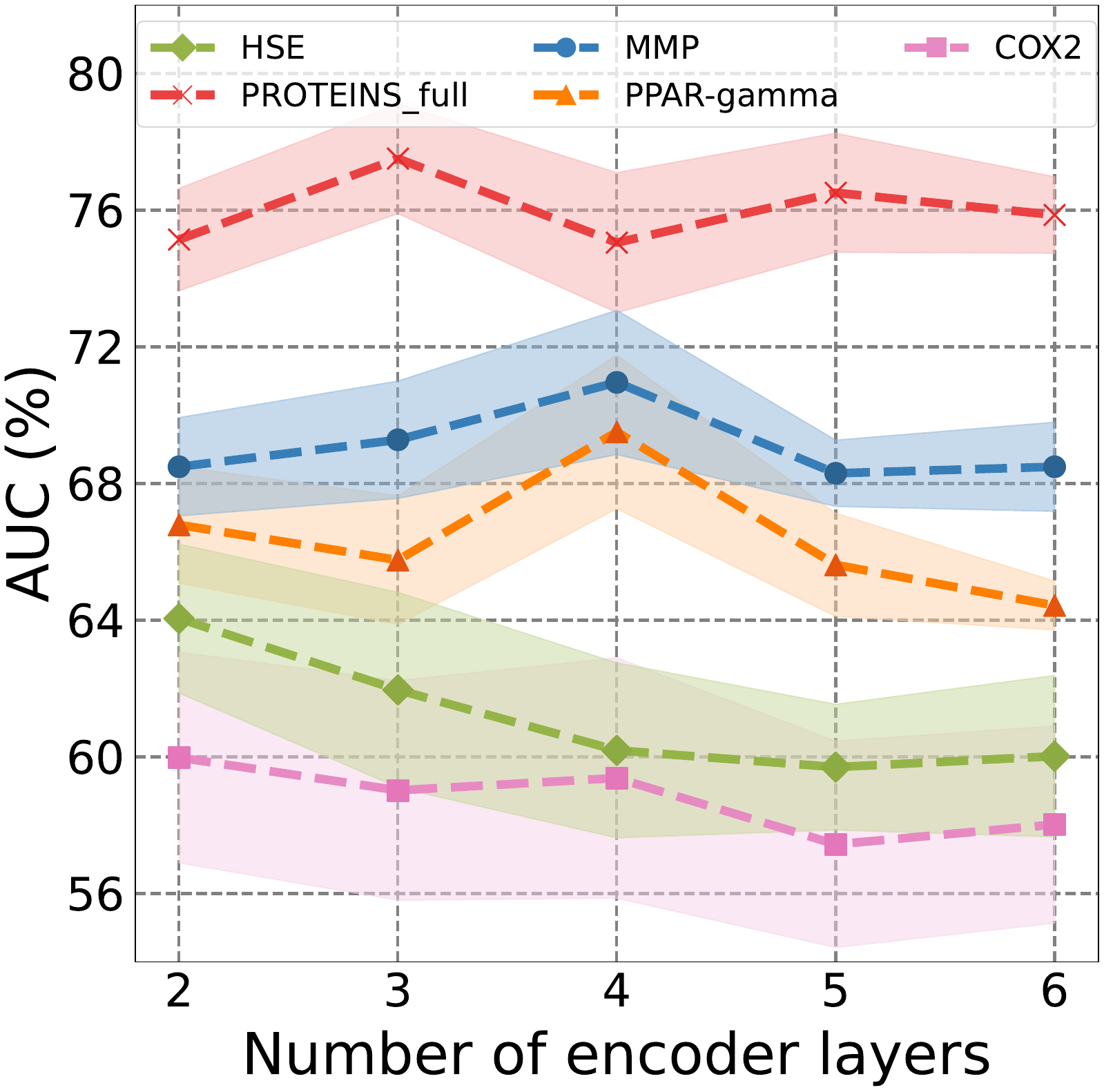}\label{fig:hyper_layers}}
    \hspace{-1mm}
    \subfigure[Different hidden dimensions.]{\includegraphics[width=0.23\textwidth]{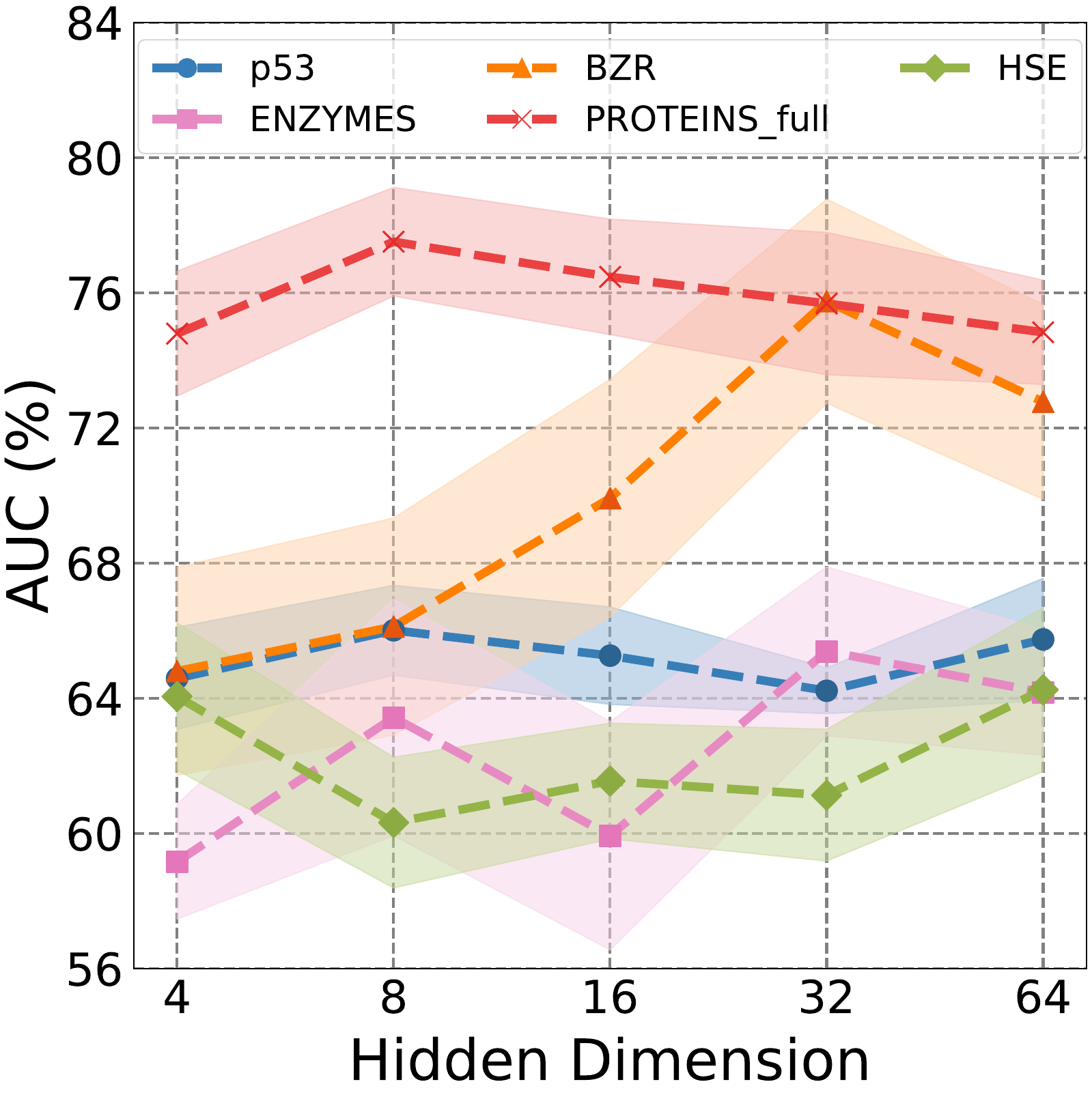}\label{fig: Hyper-parameter-HddenDimension}}
\vspace{-1em}
\caption{AUC (\%) performance w.r.t. the number of encoder layers and hidden dimensions on representative datasets.}
\label{fig: HddenDimension}
\vspace{-1em}
\end{figure}

\subsection{Ablation Study}
We conduct an ablation study on four representative datasets to investigate the effects of the two key components: hypergraph- channel and hyperbolic learning. For convenience, let \textit{w/o HyperG} and \textit{w/o HyperB} denote the customized variants of HC-GLAD without hypergraph-channel and hyperbolic learning, respectively. As shown in Figure \ref{fig: Ablation-study}, we can observe that HC-GLAD consistently achieves the best performance against two variants, demonstrating that hypergraph learning and hyperbolic learning are necessary to get the best detection performance. Compared with HC-GLAD, the poor performance of \textit{w/o HyperG} proves the importance of considering node group information and introducing hypergraph learning to this field. The poor performance of \textit{w/o HyperB} proves the importance of introducing hyperbolic learning to UGAD. Additionally, we find that on these datasets, hyperbolic learning has a more pronounced impact compared to hypergraph learning.

\subsection{Hypergraph Motifs Impact}
To further evaluate the effectiveness of using the triangular relationship as the “gold motif” for hypergraph construction, we conduct an additional experiment by modifying the motif structure. Specifically, instead of using a complete triangle, we remove one edge from the triangle as a triangle variant to analyze the impact of different motif structures. Table~\ref{motif_variant} shows a consistent decline in AUC performance across all six datasets. That indicates that a complete triangle captures essential high-order correlations between nodes, which are critical for anomaly detection. With an acceptable computational overhead, the triangle motif provides a more comprehensive and robust representation of the hypergraph.

\subsection{Hyper-parameter Analysis}
\textbf{Trade-off parameter \texorpdfstring{$\lambda_1$}{lambda}.}
In $\mathcal{L}_{total}$ in Eq.~(\ref{Equation: Loss_total}), $\lambda_1$ and $\lambda_2$ are trade-off parameters that determine weights of the graph-channel and hypergraph-channel, respectively. To investigate their impact on model performance, we conduct experiments on representative datasets as shown in Figure \ref{fig: Hyper-parameter-lambda_1}. For simplicity, we set $\lambda_2 = 1 - \lambda_1$. We observe that as $\lambda_1$ increases from 0.1 to 0.9, the performance trend varies across different datasets. However, its variation does not significantly affect model performance, indicating relative stability and high robustness of the proposed model.

\textbf{The Number of Encoder Layers.}
To investigate the impact of encoder layers on model performance, we conduct experiments on five representative datasets as shown in Figure~\ref{fig:hyper_layers}. We observe that when the number of layers is set to 3 or 4, the model exhibits promising performance. However, increasing more layers yields no significant performance improvements. When the number of layers reaches 6, a phenomenon of performance degradation commonly occurs, which we attribute to over-smoothing.

\textbf{Hidden Dimension.}
We also investigate the impact of hidden dimensions on model performance through experiments on five datasets as shown in Figure \ref{fig: Hyper-parameter-HddenDimension}. Based on our observations, we can preliminarily conclude that a higher dimension does not necessarily improve performance. In certain intervals, increasing the dimension can actually degrade the model's performance. The impact of dimension change on model performance is minimal across most datasets, with the model performance remaining relatively stable.

\begin{figure}
    \centering
    \vspace{-1em}
    \subfigure{\includegraphics[width=0.23\textwidth]{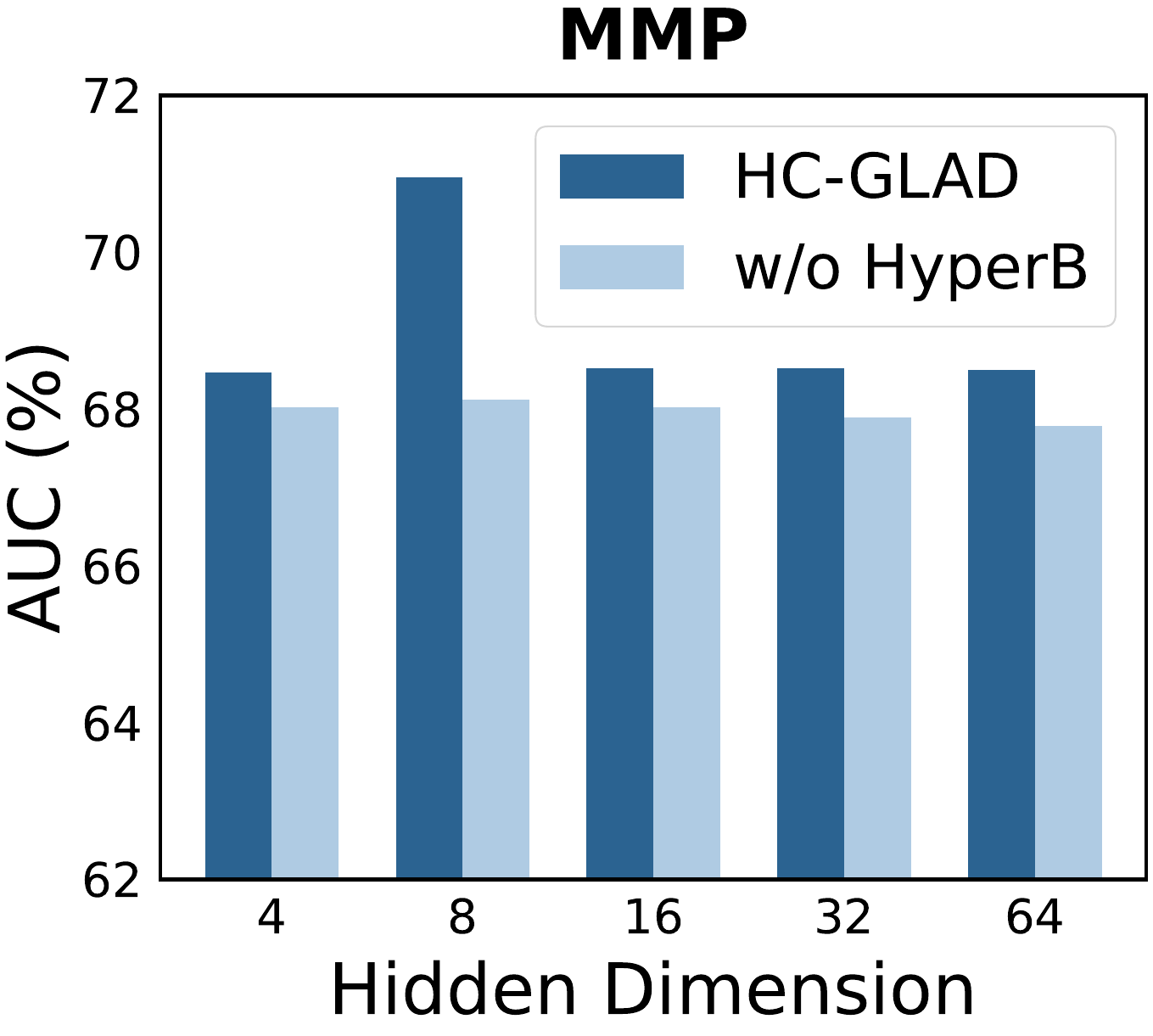}}
    \hspace{0.2mm}
    \subfigure{\includegraphics[width=0.23\textwidth]{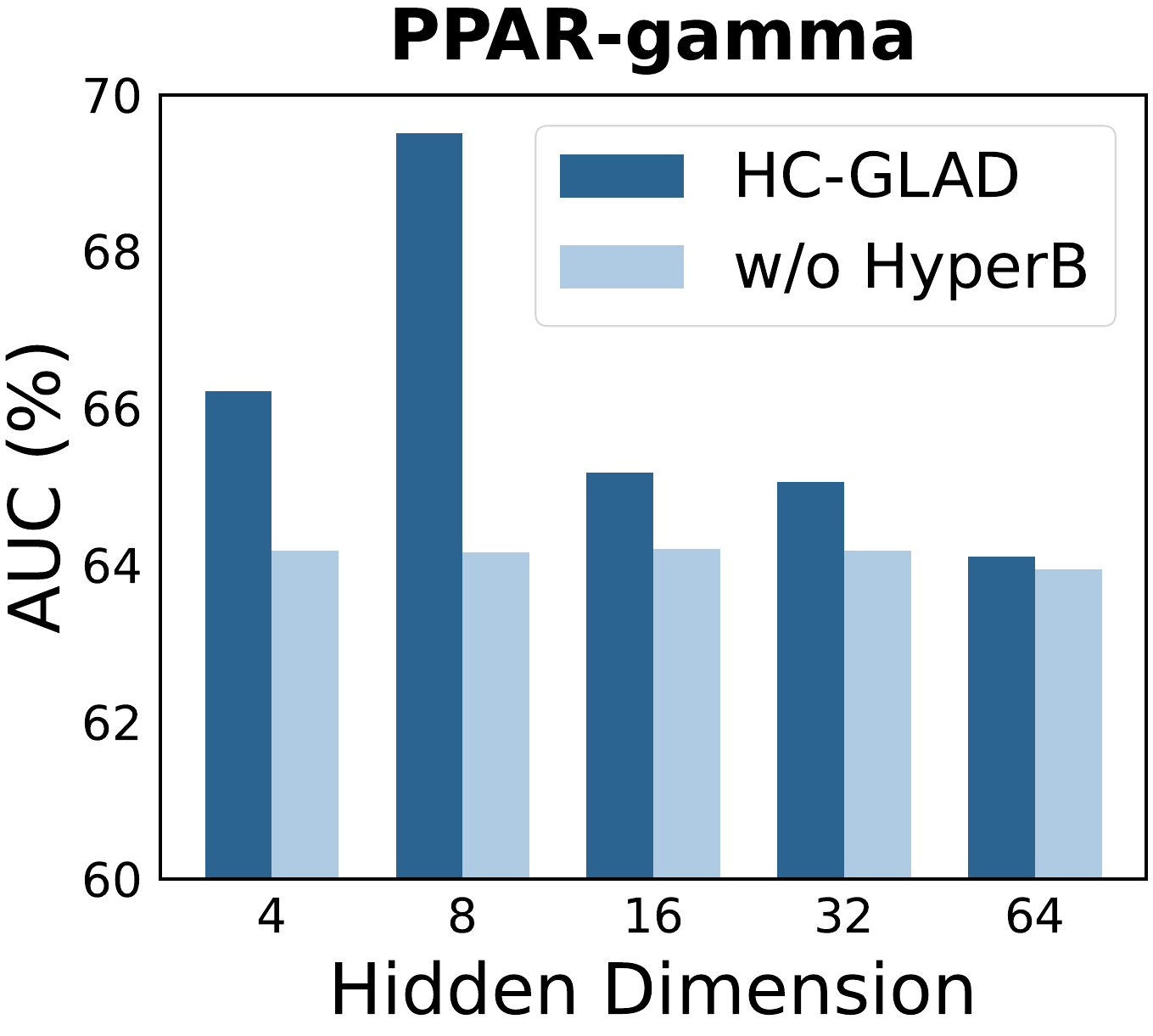}}
\vspace{-1em}
\caption{The effect of hidden dimension on HC-GLAD (hyperbolic model) and \textit{w/o HyperB} (Euclidean model) performance in terms of AUC.}
\label{fig: w/o HyperB_hidden_dimension}
\end{figure}

To further explore the impact of hidden dimensions on model performance in hyperbolic and Euclidean space, we conduct additional experiments on two representative datasets using our model HC-GLAD and its variant, \textit{w/o HyperB} (where the embedding learning is performed in Euclidean space). As shown in Figure~\ref{fig: w/o HyperB_hidden_dimension}, HC-GLAD in hyperbolic space consistently outperforms the \textit{w/o HyperB} in Euclidean space across all hidden dimensions. And the performance gap between the two models narrows with increasing hidden dimension. However, the computational overhead also grows accordingly. These findings further validate the superiority of hyperbolic space for representation.

\subsection{Visualization}
To better understand our proposed model, we employ T-SNE \cite{2008_JMLR_tSNE} to visualize the embeddings learned by HC-GLAD on AIDS dataset as shown in Figure~\ref{fig: Visualization}. 
The graph embeddings of view $v_1$ and $v_2$, learned via the graph-channel or hypergraph-channel, successfully differentiate most normal graphs from anomalous ones, demonstrating the strong representational capacity of our framework. However, certain subtle anomalies remain harder to detect, indicating that relying solely on one channel or view may still miss more nuanced distinctions.
It is ultimately the mechanism designed by HC-GLAD that distinctly differentiates normal graphs from anomalous graphs. This demonstrates the effectiveness of our scoring mechanism.

\section{Conclusion}
In this paper, we propose a novel framework named HC-GLAD, which integrates the strength of hypergraph learning and hyperbolic learning to jointly enhance the performance of UGAD. In concrete, we employ hypergraph learning built on gold motifs to exploit the node group information and utilize hyperbolic geometry to explore the latent hierarchical information. To the best of our knowledge, this is the first work to simultaneously introduce hypergraph exploiting node group information and hyperbolic geometry to the UGAD task. Through extensive experiments, we validate the superiority of HC-GLAD on 13 real-world datasets corresponding to different fields. One limitation of our method is that the integration of multiple learning paradigms in our framework may introduce increased computational cost. A more detailed time complexity analysis can be found in Appendix ~\ref{time_complexity_analysis}. In the future, we will explore the design of lightweight yet efficient frameworks to overcome this limitation.


\clearpage
\bibliographystyle{ACM-Reference-Format}
\bibliography{HC-GLAD}

\clearpage

\appendix

\section{Supplementary Related Work}
\label{appendix:gcl}
\subsection{Graph Contrastive Learning}
Graph contrastive learning employs the principle of mutual information maximization to extract rich representations by optimizing instances with similar semantic content \cite{2021_TKDE_Survey_GraphSSL,2022_TKDE_Survey_GraphSSL}. This approach has gained widespread application for achieving outstanding performance in unsupervised graph representation learning \cite{2020_ICLR_InfoGraph, 2020_NeurIPS_GraphCL, 2020_KDD_GCC, 2019_ICLR_DGI, 2021_WWW_GCL, 2023_AAAI_GREET}. For example, GraphCL \cite{2020_NeurIPS_GraphCL} proposes four types of data augmentations for graph-structured data to create pairs for contrastive learning. In the context of graph classification, InfoGraph \cite{2020_ICLR_InfoGraph} aims to maximize the mutual information between graph-level and substructure-level representations, with the latter being computed at various scales. Recent research has also applied graph contrastive learning to the field of graph-level anomaly detection. For instance, GLADC \cite{2022_ScientificReports_GLADC} captures both node-level and graph-level representations using a dual-graph encoder module within a contrastive learning framework. GOOD-D \cite{2023_WSDM_GOOD-D} detects anomalous graphs by identifying semantic inconsistencies across different granularities through a hierarchical contrastive learning framework. CVTGAD \cite{2023_ECMLPKDD_CVTGAD} similarly incorporates graph contrastive learning principles, utilizing transformer for unsupervised graph anomaly detection and explicitly accounting for co-occurrence between different views.

\section{Supplement of Experiments}
\subsection{Datasets}
\label{appendix:datasets_details}
More details about datasets we employed in our experiments are illustrated in Table \ref{datasets}.

\begin{figure}[]
    \centering
    \subfigure{\includegraphics[width=0.23\textwidth]{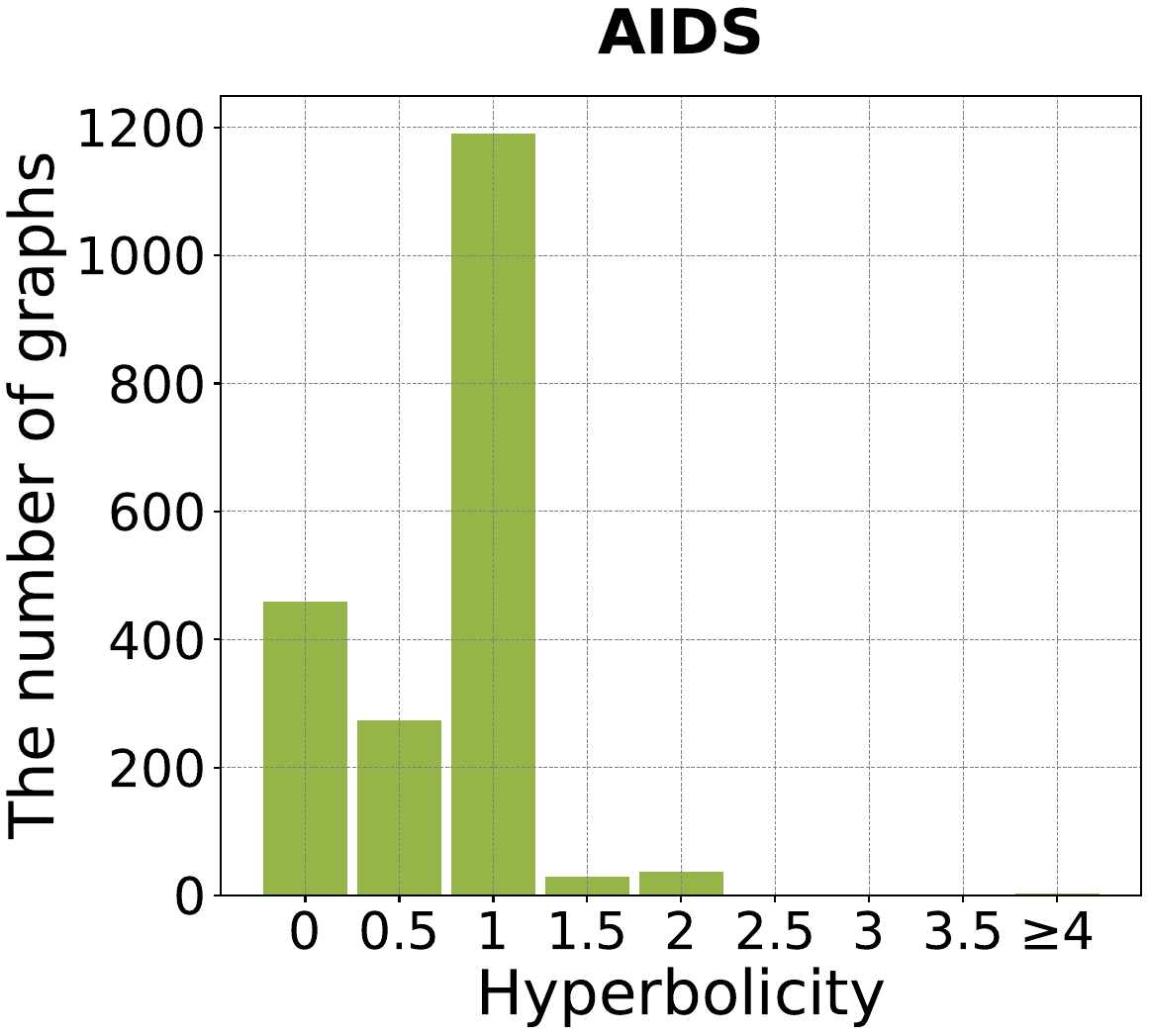}}
    \hfill
    \subfigure{\includegraphics[width=0.23\textwidth]{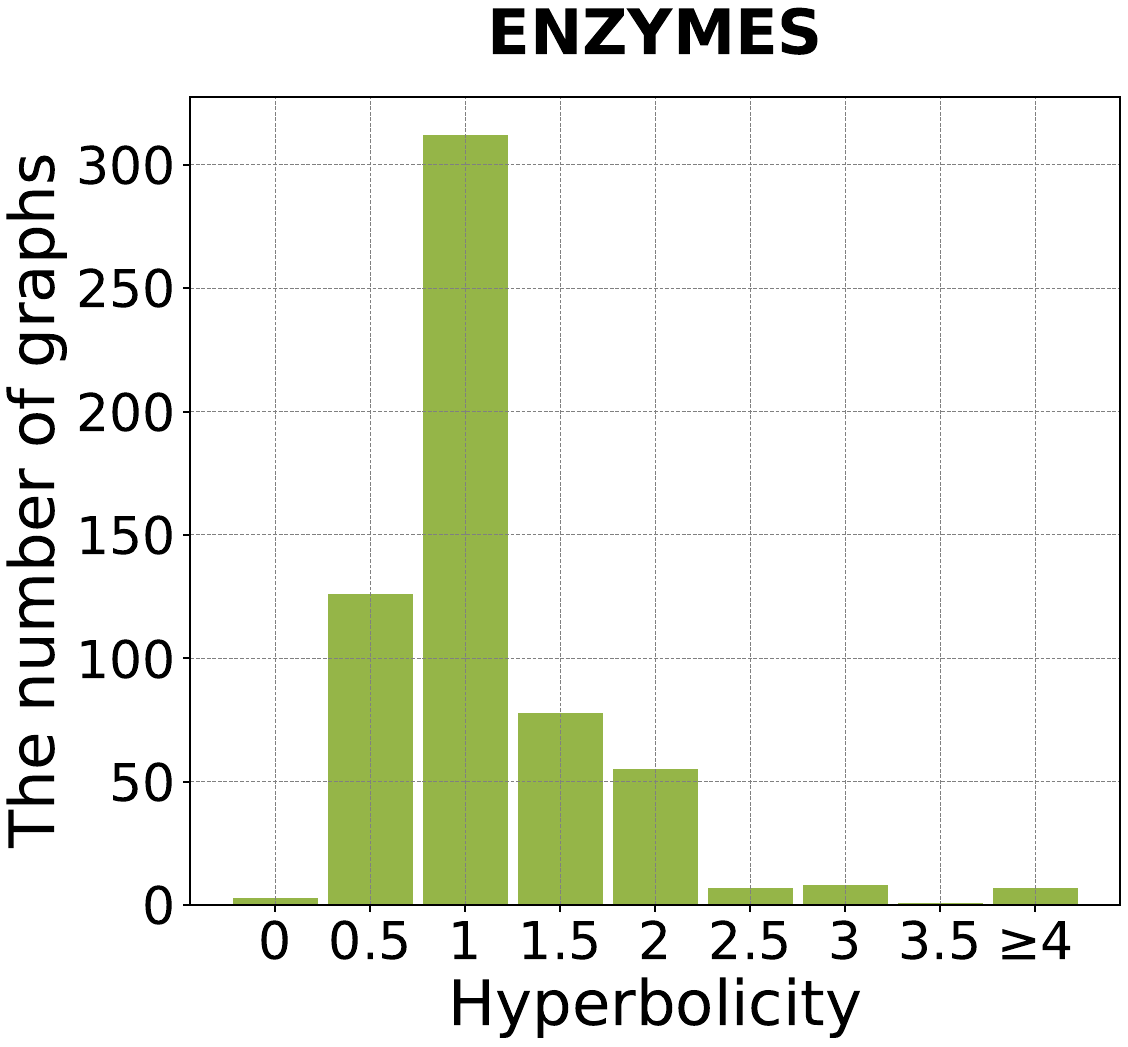}}
    \subfigure{\includegraphics[width=0.23\textwidth]{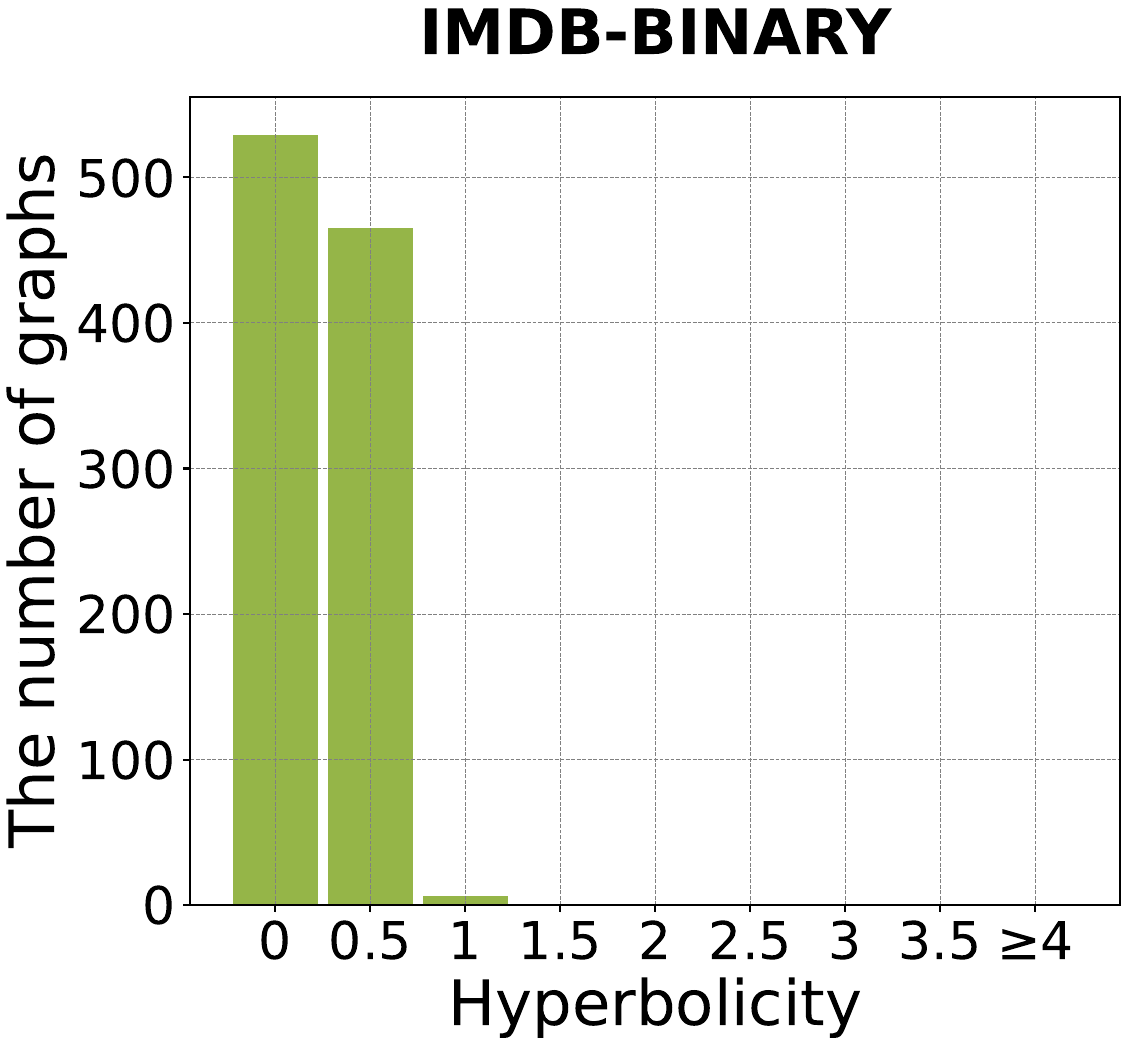}}
    \hfill
    \subfigure{\includegraphics[width=0.23\textwidth]{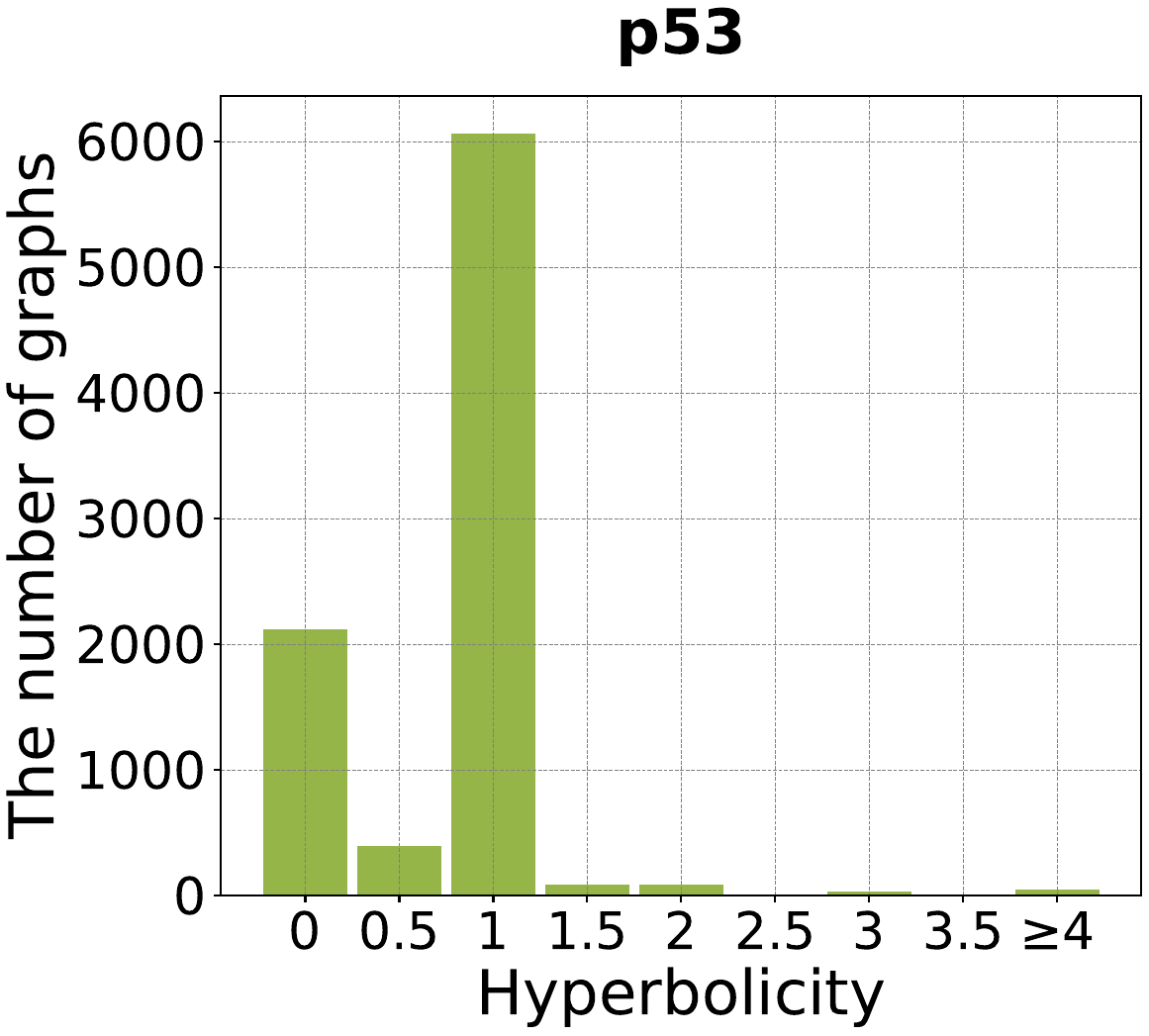}}
    \subfigure{\includegraphics[width=0.23\textwidth]{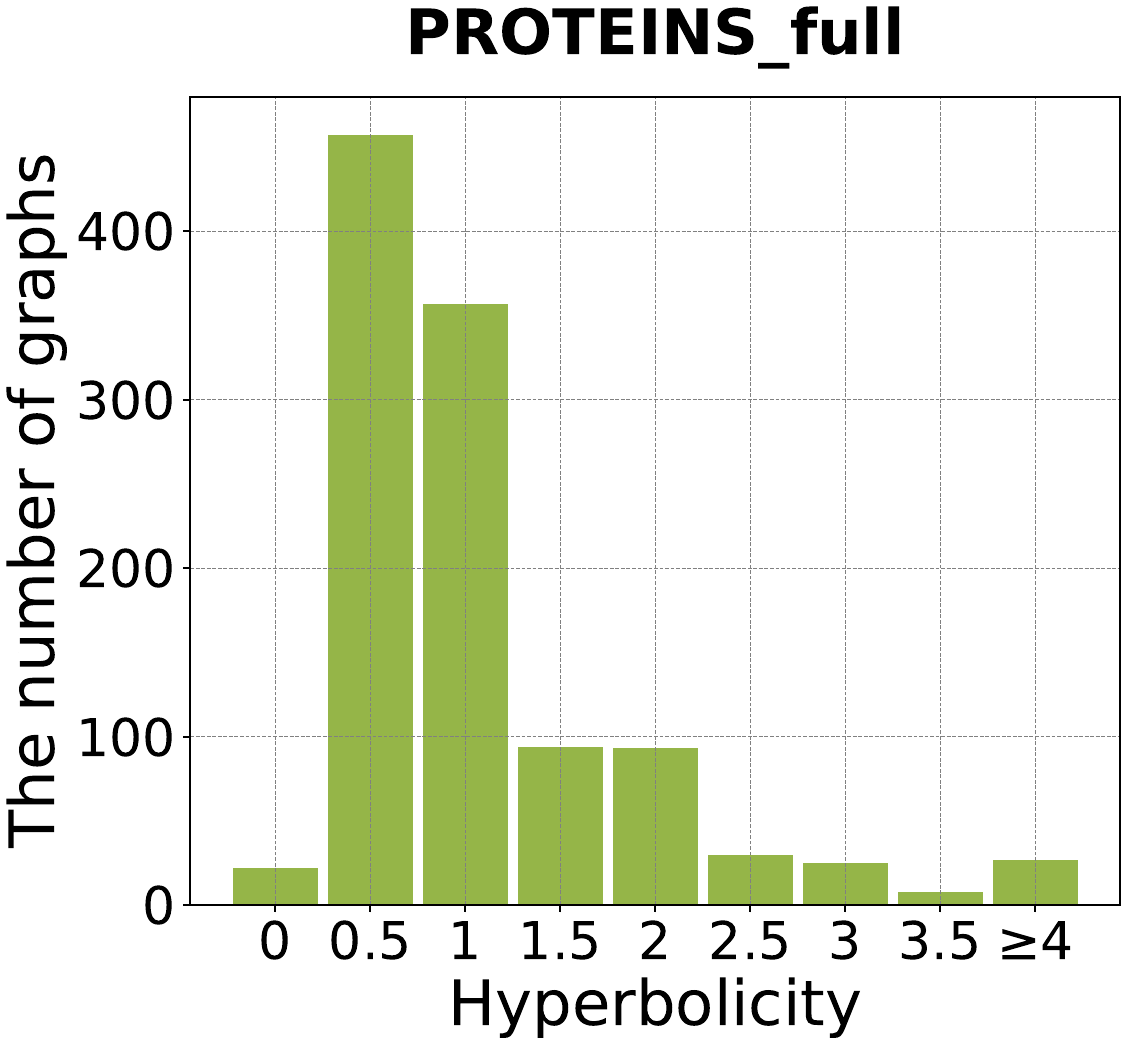}}
    \subfigure{\includegraphics[width=0.23\textwidth]{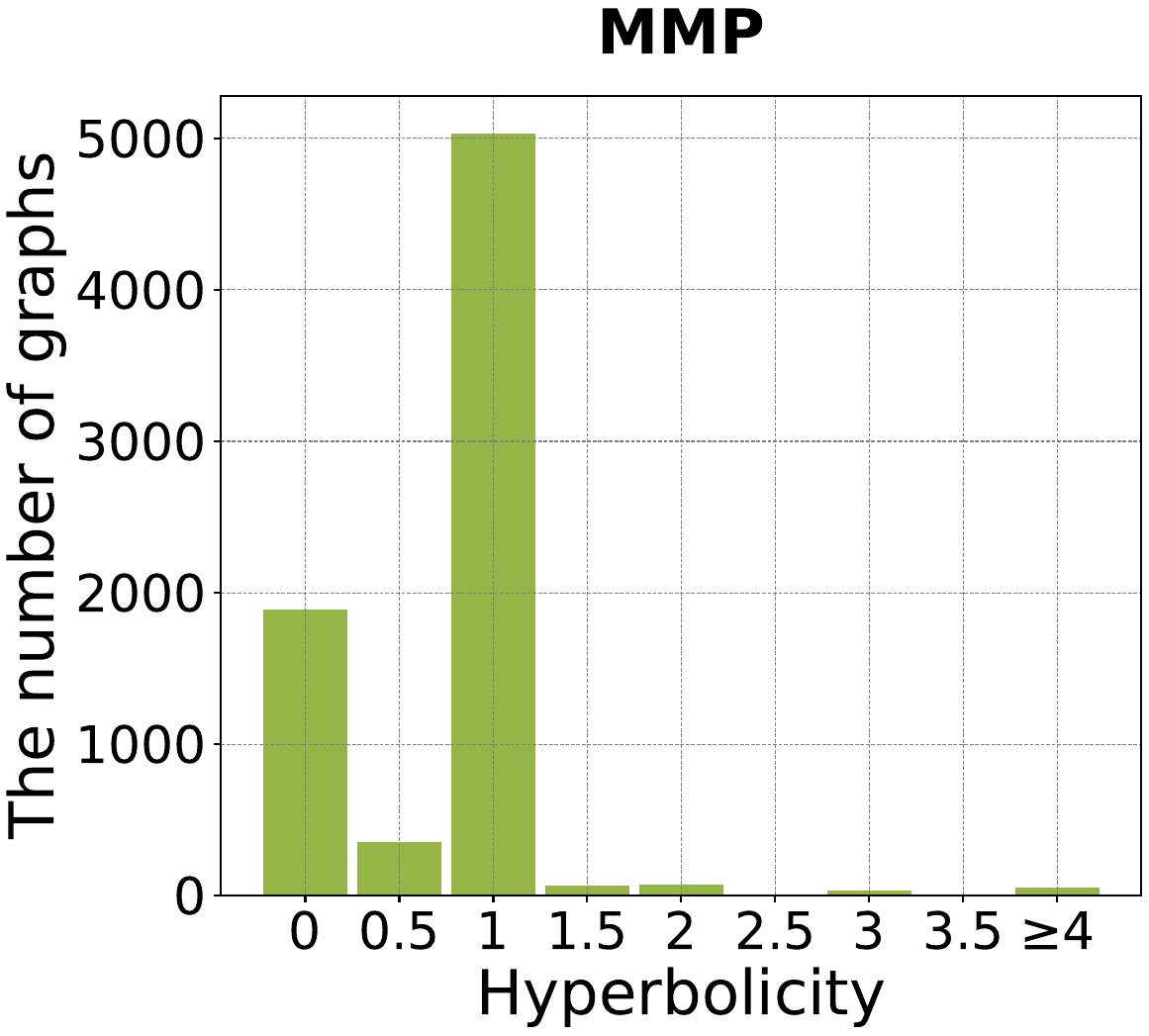}}
\vspace{-0.9em}
\caption{Hyperbolicity distribution in detail on representative datasets where the X-axis represents the hyperbolicity of the graph and the Y-axis represents the number of graphs with corresponding hyperbolicity values on the X-axis.}
\label{fig:dataset_hyp}
\end{figure}

\subsection{Hyperbolicity}
\label{appendix:hyperbolic_definition}
The hyperbolicity is based on the 4-node condition, a quadruple of distinct nodes $n_1$, $n_2$, $n_3$, $n_4$ in a graph. Let $\pi = (\pi_1, \pi_2, \pi_3, \pi_4)$ be a permutation of node indices 1, 2, 3, and 4, such that

\begin{equation}
\begin{split}
    {S}_{n_{1}, n_{2}, n_{3}, n_{4}} &= d(n_{\pi_{1}}, n_{\pi_{2}}) + d(n_{\pi_{3}}, n_{\pi_{4}}) \\
    &\leq {M}_{n_{1}, n_{2}, n_{3}, n_{4}} = d(n_{\pi_{1}}, n_{\pi_{3}}) + d(n_{\pi_{2}}, n_{\pi_{4}}) \\
    &\leq {L}_{n_{1}, n_{2}, n_{3}, n_{4}} = d(n_{\pi_{1}}, n_{\pi_{4}}) + d(n_{\pi_{2}}, n_{\pi_{3}}), 
\label{hyperbolicity_1}
\end{split}
\end{equation}

where $d$ is the shortest path length, and define
\begin{equation}
    \delta^{+} = \frac{{L}_{n_{1}, n_{2}, n_{3}, n_{4}} - {M}_{n_{1}, n_{2}, n_{3}, n_{4}}}{2}.
\label{hyperbolicity_2}
\end{equation}

The worst-case hyperbolicity \cite{1987_Hyperbolic_Groups} is defined as the maximum value of $\delta$$^{+}$ among all quadruples in the graph, i.e.,
\begin{equation}
    \delta_{worst}= \max_{n_1, n_2, n_3, n_4} \{\delta^+\}.
\end{equation}

The average hyperbolicity \cite{2014_define_avg_hyperbolicity} is defined as the average value of $\delta$$^{+}$ among all quadruples in the graph, i.e.,
\begin{equation}
    \delta_{avg}= \frac{1}{\binom{n}{4}} \sum_{n_1, n_2, n_3, n_4} \{\delta^+\},
\end{equation}

where $n$ is the number of nodes in the graph.

A graph $\mathcal{G}$ is called $\delta$-hyperbolic if $\delta_{worst}(\mathcal{G}) \leq \delta$ 
\cite{2014_define_avg_hyperbolicity}. We adopt the aforementioned $\delta_{worst}$ as the hyperbolicity $\delta$ of the datasets, which to some extent reflects the underlying hyperbolic geometry of the graph.

\begin{table*}[]
\caption{The statistics of datasets of our experiments from TUDataset \cite{2020_arXiv_TuDataset}.}
\vspace{-0.5em}
\label{datasets}
\centering
\fontsize{7}{8}\selectfont
    \renewcommand{\arraystretch}{2}
    \begin{tabular}{c|ccccccccccccccc}
    \toprule
    \textbf{Dataset}    & \textbf{PROTEINS\_full} & \textbf{ENZYMES} & \textbf{AIDS} & \textbf{DHFR} & \textbf{BZR} & \textbf{COX2} & \textbf{DD} & \textbf{REDDIT-B} & \textbf{HSE} & \textbf{MMP} & \textbf{p53} & \textbf{PPAR-gamma} & \textbf{IMDB-B} \\ \midrule
    \textbf{Graphs}     & 1113     & 600      & 2000     & 467      & 405     & 467      & 1178      & 2000      & 8417      & 7558      & 8903     & 8451     & 1000    \\ 
    \textbf{Avg. Nodes} & 39.06    & 32.63    & 15.69    & 42.43    & 35.75   & 41.22    & 284.32    & 429.63    & 16.89     & 17.62     & 17.92    & 17.38    & 19.77   \\ 
    \textbf{Avg. Edges} & 72.82    & 62.14    & 16.20    & 44.54    & 38.36   & 43.45    & 715.66    & 497.75    & 17.23     & 17.98     & 18.34    & 17.72    & 96.53   \\ 
    \bottomrule
    \end{tabular}
\end{table*}

\subsection{Time Complexity Analysis}
\label{time_complexity_analysis} 
While data augmentation follows the same standard process as~\cite{2023_WSDM_GOOD-D}, the hypergraph construction, whose core calculation is the Eq.~(\ref{Eq: motifs}), can be efficiently computed using sparse matrices \cite{2018_AAAI_MPR, 2021_WWW_MHCN}.
The data augmentation and hypergraph construction are both performed once during the preprocessing stage, and the time complexity of our model is mainly in the encoder and loss term. Let $M$ denote the number of hyperedges, $e$ denote the average number of edges, $n$ denotes the average number of nodes, $m$ denote the total number of graphs, $\mathcal{B}$ denote the batch size, $d$ denote latent embedding dimension, $d_{\text{in}}$ denote the dimension of input-layer embedding and L denote the number of encoder layers. For the graph encoder GNN, the time complexity is $\mathcal{O}(mLed + mLnd^{2} + mndd_{\text{in}})$. For the hypergraph encoder HGCN, the time complexity is $\mathcal{O}(mLMd + mLnd^{2} + mMdd_{\text{in}})$. For node-level loss, the time complexity is $\mathcal{O}(mn^{2}d)$. For graph-level loss, the time complexity is $\mathcal{O}(m\mathcal{B}d)$. So our model's time complexity is $\mathcal{O}(mLd(e + nd + M) + md(nd_{\text{in}} + Md_{\text{in}} + n^{2} + \mathcal{B}))$ for training, ignoring constant coefficient term and the smaller terms.

\end{document}